\documentclass[pdflatex,sn-mathphys]{sn-jnl}

\usepackage{amsmath, amssymb} 
\usepackage{amsthm}
\usepackage{float}
\usepackage{hyperref}
\usepackage{mathtools}
\usepackage{subcaption}
\usepackage{tablefootnote}

\jyear{2022}%

\newtheoremstyle{mytheoremstyle}{8pt}{8pt}{\itshape}{}{\bfseries}{.}{.5em}{} 

\theoremstyle{mytheoremstyle}

\newtheorem{hyp}{Hypothesis}
\theoremstyle{thmstyletwo}%

\theoremstyle{thmstylethree}%
\theoremstyle{mytheoremstyle}
\newtheorem{definition}{Definition}%

\DeclarePairedDelimiter\abs{\lvert}{\rvert}%

\begin{document}
\noindent

\title[Can recurrent neural networks learn process model structure?]{Can recurrent neural networks learn process model structure?}


\author[1]{\fnm{Jari} \sur{Peeperkorn}}\email{jari.peeperkorn@kuleuven.be}

\author[2,1]{\fnm{Seppe} \sur{vanden Broucke}}\email{seppe.vandenbroucke@kuleuven.be}

\author[1]{\fnm{Jochen} \sur{De Weerdt}}\email{jochen.deweerdt@kuleuven.be}

\affil[1]{\orgdiv{Research Center for Information Systems Engineering (LIRIS)}, \orgname{KU Leuven}, \orgaddress{\city{Leuven}, \country{Belgium}}}

\affil[2]{\orgdiv{Department of Business Informatics and Operations Management}, \orgname{Ghent University}, \orgaddress{ \city{Ghent}, \country{Belgium}}}



\abstract{
Various methods using machine and deep learning have been proposed to tackle different tasks in predictive process monitoring, forecasting for an ongoing case e.g. the most likely next event or suffix, its remaining time, or an outcome-related variable. Recurrent neural networks (RNNs), and more specifically long short-term memory nets (LSTMs), stand out in terms of popularity. In this work, we investigate the capabilities of such an LSTM to actually learn the underlying process model structure of an event log. We introduce an evaluation framework that combines variant-based resampling and custom metrics for fitness, precision and generalization. We evaluate 4 hypotheses concerning the learning capabilities of LSTMs, the effect of overfitting countermeasures, the level of incompleteness in the training set and the level of parallelism in the underlying process model. We confirm that LSTMs can struggle to learn process model structure, even with simplistic process data and in a very lenient setup. Taking the correct anti-overfitting measures can alleviate the problem. However these measures did not present themselves to be optimal when selecting hyperparameters purely on predicting accuracy. We also found that decreasing the amount of information seen by the LSTM during training, causes a sharp drop in generalization and precision scores. In our experiments, we could not identify a relationship between the extent of parallelism in the model and the generalization capability, but they do indicate that the process' complexity might have impact. 
}




\maketitle

\section{Introduction}\label{sec1}

Predictive process monitoring refers to the prediction of future characteristics of process instances. In recent years, a variety of solutions has been proposed for addressing different prediction tasks, in particular predicting the most likely next event or suffix of a case, its remaining time, or an outcome-related variable. Many of these solutions rely on deep learning, with Recurrent Neural Networks (RNNs) being the most popular architecture. More specifically, one type of RNN has received a majority of the attention: Long Short-Term Memory neural networks (LSTMs). However, to the best of our knowledge, no research has investigated whether LSTMs actually learn process model structure. This characteristic is never explicitly evaluated given the focus on the dedicated prediction tasks listed above. Nevertheless, it is a common assumption in process mining that discovered models, whether used in a predictive setting or not, are capable of generalizing the observed behaviour into process model structures. Moreover, one can certainly question the predictive qualities of deep learning models if they actually fail to generalize control-flow behaviour, given that they might be memorizing training data instead of truly learning. Therefore, the key question becomes to what extent deep learning-based models such as LSTMs are able to learn and generalize control-flow behaviour from a possibly incomplete set of examples. Given the fact that often processes can display a wide arrange of behavior, resulting in often incomplete event logs ~\cite{reallife}, this is a key issue of predictive process monitoring. 

In previous work, we have proposed a framework to assess a next event prediction model's capability to generalize and understand process model structure~\cite{Jari}. Based on simulation and a variant-based resampling procedure, metrics were defined to measure fitness, precision and generalization of the predictive model in function of its model learning capacity. A limited experiment was presented uncovering possible generalization issues. In this paper, we extend our previous work in three ways. First, we extend the framework with an additional metric type, which allows for different levels of stringency. Second, we extend the experimental setup by contrasting two different approaches for optimizing hyperparamaters: one approach based on a post-hoc optimization in function of the metrics of our framework, and a second approach with more realistic hyperparameters obtained by optimizing model accuracy for next event prediction. Finally, based on these extensive experiments, we show that process model structure learning capabilities of RNN-based neural networks in predictive process monitoring is not guaranteed. By relating this observation to event log and process model characteristics, we can better understand and position issues, which in turn allows us to propose several ideas for further research. 


In summary, the main question addressed in this paper can be stated as follows: to which degree can LSTM neural networks, trained for next event prediction, learn process model structure? In order to establish a set of discernible experiments, we partition this question into four key hypotheses:

\begin{hyp}[H\ref{hyp:1}] \label{hyp:1}
LSTMs built for next event prediction do not properly learn process model structure.
\end{hyp}
\begin{hyp}[H\ref{hyp:2}] \label{hyp:2}
The incorporation of particular anti-overfitting measures can allow an LSTM to learn process model structure.
\end{hyp}
\begin{hyp}[H\ref{hyp:3}] \label{hyp:3}
The higher the degree of incompleteness of the training set in terms of control-variants observed, the worse an LSTM will generalize the control-flow behavior.
\end{hyp}
\begin{hyp}[H\ref{hyp:4}] \label{hyp:4}
The more parallel behavior in a process, the worse an LSTMs’ generalization capability.
\end{hyp}

By addressing these questions, we hope to provide insights that could inspire future work on the improvement of predictive process mining models. The rest of this paper is organized as follows. First, Section \ref{preliminaries} introduces essential notation and a discussion of Recurrent Neural Networks, followed by Section \ref{Related} discussing some relevant related work. Next, in Section \ref{Metrics}, we introduce the metrics for measuring fitness, precision and generalization. In Section \ref{Setup} we describe the experimental setup with which we tested the hypotheses mentioned above, by first introducing the selected artificial process models, followed by a discussion of the hyperparameter search and the experiments itself. The results of these experiments are shown and discussed in Section \ref{Results}, before concluding the paper in Section \ref{Conclusion}. 
The data and code used in this work can be found online\footnote{\url{https://github.com/jaripeeperkorn/LSTM_Process_Model_Structure}}.

\section{Preliminaries}
\label{preliminaries}
\subsection{Notation}

In this section we introduce the preliminary concepts required to explain our approach. The main data sources used in predictive process monitoring are \textit{event logs}, which record the executions of different activities within a business process. A \textit{case} or \textit{process instance}, with its corresponding \textit{case id}, refers to one specific execution of this process. Each \textit{event} has a \textit{case id} specifying to which case it belongs, an \textit{activity name} (or \textit{event class}) indicating which type of activity was executed, and a \textit{timestamp} indicating the completion time of the activity. An \textit{event} can optionally contain other \textit{case} or \textit{event attributes} as well. An \textit{event} can then be formally defined as: 
 \begin{definition}[Event] An \textit{event} is a tuple $\left(a, c, t, (d_1, v_1),\dots,(d_m, v_m)\right)$ where $a$ is an activity class, c is a case id, t is a timestamp and $(d_1, v_1),\dots,(d_m, v_m)$ (with $m\geq0$) are possible case or event attributes and their values.
 \end{definition}
 The complete sequence of events recorded for a given \textit{case} forms a \textit{trace}.
 \begin{definition}[Trace] A \textit{trace} is a non-empty sequence $\sigma = [e_1,\dots,e_n]$ of events such that $\forall i\in[1 \dots n], e_i\in\mathcal{E}$ (the set of possible event classes) and $\forall i,j \in[1\dots n]$ $e_i.c = e_j.c$, i.e. all events in the trace refer to the same case id.  
 \end{definition}
 An \textit{event log} can then be defined as a multiset of \textit{traces} recording executions of completed cases. A \textit{simple event log} can be defined as a set of \textit{simple traces}, sequences of only the activity labels without timestamps and other possible attributes, but maintaining the order of events. This basically corresponds to a set of sequences with only activity labels. In the rest of this work, only \textit{simple event logs} and \textit{simple traces} are used, and the term \textit{simple} is dropped. We use $\abs{L}$ to denote the amount of traces in \textit{event log L}. The set of all possible different \textit{(simple) traces} recorded by executing the process are called the \textit{control-flow variants} or \textit{variants} in short. Note that we use variant to depict \emph{trace} variants, i.e. different variations of executions of the same process, and not \emph{process} variants (depicting different processes). Formally:
 \begin{definition}[Control-flow variant, $Var$] Given a simple event log $L$, there exists a set of unique simple traces $Var(L)$ (the control flow variants) such that: \\
  $\forall$ $\sigma=[a_1,\dots,a_n] \in L:$ $\exists!$ $v=[v_1,\dots,v_n] \in Var(L)$ s.t. $\forall i\in[1\dots n]: a_i = v_i $
 \end{definition}

Accordingly we define $Occ(v, L)$ as a function which returns the amount of times control-flow variant $v$ occurs in $L$. 
\begin{definition}[\textit{Occ}] Given a simple event log $L$ and $v \in Var(L)$:\\
$Occ(v, L)$ is equal multiplicity of $v$ in multiset $L$.
\end{definition}

 The predictive models assessed in this work, use \textit{prefixes} as input. A \textit{prefix} consists of the first $l$ events of a complete \textit{trace}.
 \begin{definition}[Prefix] Given a trace $\sigma = [e_1,\dots,e_n]$ and a positive integer $l\leq n$, $\text{\textit{prefix}}\left(\sigma, l\right) = [e_1,\dots,e_l]$.
\end{definition}
 Because machine learning models cannot distinguish different categories without some sort of encoding, we also define:
 \begin{definition}[One-hot encoding] Given the set of possible event classes $\mathcal{E}$ and one specific activity label corresponding to the $i$th activity in $\mathcal{E}$, the one-hot representation of this activity is a sparse $\abs{\mathcal{E}}$ dimensional vector, with the only non-zero entry located at position i. 
 \end{definition}
A process model is used to denote a graphical representation of a process. These models can be expressed using various different modelling languages. In this work, Petri nets~\cite{Petri}, or more specifically workflow nets, are used to visualise and play-out different artificially designed process models. Process trees are used as well, to randomly generate new artificial process models~\cite{ProcessTree}.

\subsection{Recurrent Neural Networks}

\begin{figure}[ht]
 \centerline{\includegraphics[width=0.5\linewidth]{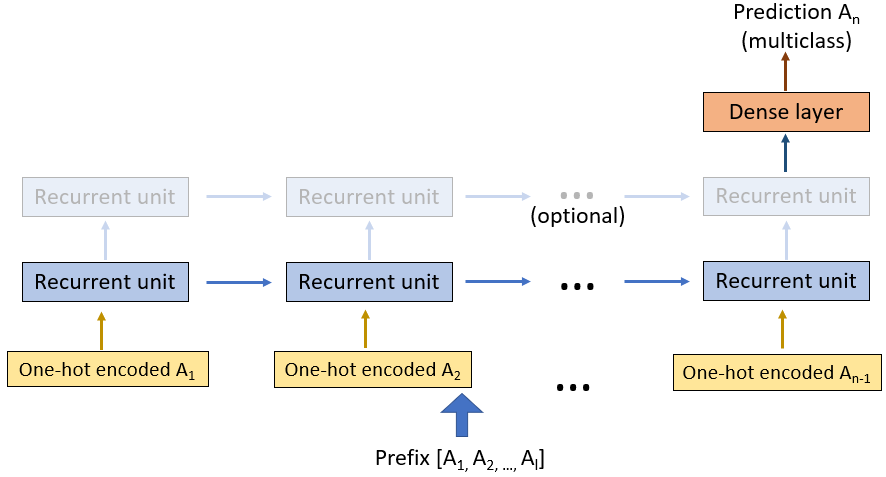}}
 \caption{An overview of the Recurrent Neural Network structure used for next event prediction.}
 \vspace{-6mm}
 \label{fig:RNN}
 \end{figure}

 Recurrent Neural Networks (RNNs) are a type of artificial neural networks suited and designed to handle sequential data. RNNs are often described as a chain of multiple feedforward neural networks, one for each time step. The recurrent (hidden) layer runs trough all of these time steps and passes on a message throughout time steps. When two layers are used, with one for each direction, such models are referred to as a bidirectional recurrent layer. An RNN can have multiple recurrent layers stacked on top of each other: within this layer the time steps are connected and at each time step the input comes from the output of the specific time step in the previous recurrent layer. Weights of RNNs are usually trained using \textit{backpropagation through time}, where the backpropagation is performed from output to input, through the hidden layers, and backwards in time. Simple RNNs have been shown to have difficulties when handling long-term dependencies, due to vanishing and exploding gradients problems. Multiple solutions, like Long Short-term Memory (LSTM)~\cite{LSTM} and Gated Recurrent Units (GRU)~\cite{GRU}, have been proposed to solve these issues throughout the years. The RNN architecture used in this paper can be found in Figure~\ref{fig:RNN}.

 The input for this model is a (simple) case prefix. The activities are presented one by one to the network, in a one-hot encoded way. One design choice omitted here, is the possible addition of an embedding layer, which converts the sparse higher-dimensional one-hot representation to a lower dimensional vector representation, i.e. a mapping trained together with the rest of the network in such a way that it is meaningful in the sense of constructing a lower-dimensional vector, which retains as much of the original topology of the input space as possible to succeed on the downstream task (next event prediction in our case). In the context of simple processes with small activity vocabularies, using embeddings is not a strict necessity. In previous work~\cite{Jari}, the inclusion or omission of an embedding layer did not seem to have a significant effect in the hyperparameter search, therefore it was decided to exclude embedding layers in this work. However, it is pointed out that vector embeddings can be beneficial with growing vocabulary sizes, which is not the case here. The one-hot encoded activity is then presented to the recurrent layer (an LSTM layer). Depending on the setting, the size (number of units in a layer) and the amount of recurrent layers will differ. We will experiment with using both uni- as well as bidirectional LSTMs. The output of the last time step of the last recurrent layer is then put through a dense layer, which will predict the next activity. It does this by providing a multiclass vector output, with for each activity from the vocabulary a probability that this would be the next event. The activity labels' position in the vector correspond to the one-hot encoding. The input sequence length was chosen for each model independently. Longer prefixes are left-truncated and shorter prefixes pre-padded (i.e. vectors with only zero's added to the start of the prefix). For the models for which we know we can do so without information loss, we opted to use shorter input sequence lengths to limit computational resources. For other models, we use the maximum trace length (minus one) as the input sequence length.

\section{Related Work}\label{Related}

In recent years, within the field of Predictive Process Monitoring (PPM), focus seems to have shifted given that a plethora of deep learning-based solutions have been proposed~\cite{Tax_2017, Evermann_2017, Mehdiyev_2017, Camargo_2019, Lin_2019, Pasquadibisceglie_2019, Taymouri_2020, Bukhsh_2021}. Because the models in this work are trained optimizing the next event prediction problem, we focus in this section on those PPM works addressing the same challenge. RNNs have been introduced in the PPM literature by Tax et al.~\cite{Tax_2017}, who proposed to use an LSTM to predict next events and corresponding timestamps in the same model, thereby relying on one-hot encoding of the activity labels as input for the LSTM, together with their timestamps. In a similar work, Evermann et al.~\cite{Evermann_2017} proposed the use of LSTM networks, specifically to predict full case suffixes, including attributes such as resources. They were the first to introduce lower-dimensional vector embeddings. Multistage approaches have also been introduced, e.g. by Mehdiyev et al.~\cite{Mehdiyev_2017}, in which the use of n-gram representations, stacked with autoencoders and deep feedforward neural network classifiers, are created for the next activity prediction task. Combining some of ideas from earlier works, Camargo et al.~\cite{Camargo_2019} use separately trained embeddings of categorical variables, together with the activities' timestamps, in order to predict both next events as well as their corresponding (future) timestamps. In addition, LSTM encoder-decoders have been experimented with by Lin et al.~\cite{Lin_2019}, using both control-flow information as well as other event attributes. 

Other deep learning solutions, beyond RNNs, have been proposed as well. Pasquadibisceglie et al.~\cite{Pasquadibisceglie_2019} introduced the use of a Convolutional Neural Network (CNN) approach to PPM. Furthermore, promising results, especially with respect to generation and generalization, were reported by Taymouri et al.~\cite{Taymouri_2020}, in which a Generative Adversarial Network approach to the problem of next event, suffix and timestamp prediction was introduced. Bukhsh~\cite{Bukhsh_2021} pioneered with the application of transformer networks for PPM. For a most recent interdisciplinary overview of both classical statistical as well as machine learning-based approaches for next element prediction, we refer to ~\cite{Tax_2020}. In an unsupervised way, Deep-TRace2Vec is a deep learning-based approach which tries to encode process behavior as a vectorial representation (embedding) of the traces~\cite{multiperspect}. The obtained output embedding could be combined with different predictive models.

Only indirect prior work exists regarding the assessment of whether RNN-based architectures trained on next event prediction can actually learn process model structure. In~\cite{Reliable}, the influence of process structure on predictions is discussed. Furthermore, ~\cite{generalizing} and ~\cite{Tax_2020} discuss the performance on a hold-out test set, from the perspective of generalization, both in process discovery and (deep learning) sequence models, from which could be concluded that LSTMs should generalize towards previously unseen traces better than existing process discovery techniques. Despite these works, generalization power of neural network models is still unclear. While it has been proven that RNNs are universal approximators~\cite{UniversalApproximators} and Turing Complete (i.e. for any given computable function, there exists a finite RNN to compute it)~\cite{siegelmann1995computational} for a while now, there has still been a lot of interest into understanding how RNNs can learn different kinds of functions. While the research question concerning the capabilities of RNNs to learn formal grammars has not been studied in the Business Process Management field extensively, still relevant related work can be found in other fields, e.g. grammar inference~\cite{Lawrence, sennhauser, Schmidhuber}. Other literature in context-free grammars discusses to which extent LSTMs are able to learn hierarchical structures (and in this way generalize)~\cite{contextfree}.  In the PPM context, it is also unclear how these deep learning models handle possible data incompleteness of the event log as is the case with real-life business process event logs~\cite{reallife}. Other work has shown that when applying explainability metrics on LSTMs trained in a PPM context, the output provided is not always sensible (due to its increased complexity)~\cite{Stevens2021}. This strengthens the need for research exploring the exact learning capabilities of LSTM models when trained on process data.

\section{A framework for measuring process model structure learning}\label{Metrics}

\subsection{Framework}
\begin{figure}[ht]
 \centerline{\includegraphics[width=0.99\linewidth]{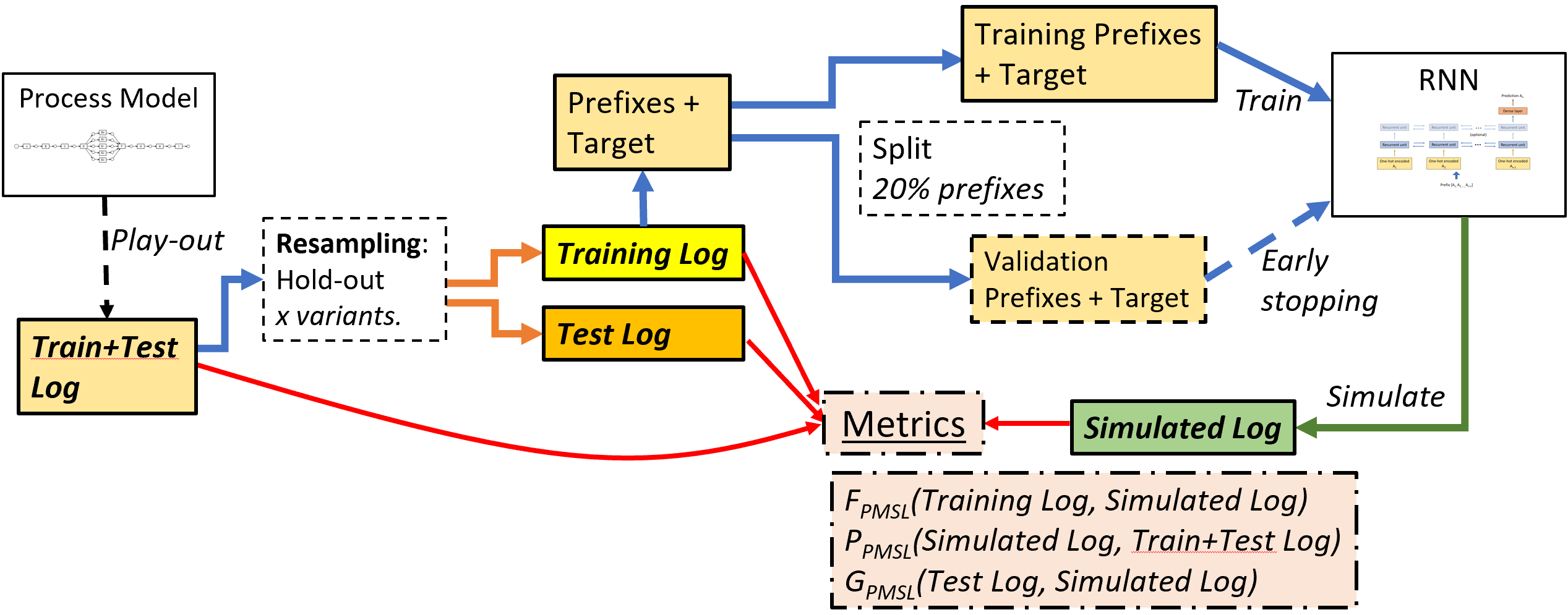}}
 \caption{The framework used to test the process model structure learning capabilities of a predictive model, in this case an RNN. It is based on a variant-based sampling scheme, and uses a process model as input.}
 \vspace{-6mm}
 \label{fig:Setup}
 \end{figure}

In order to investigate the process structure learning capabilities of a predictive algorithm, we have designed a framework, as shown in \ref{fig:Setup}. 

The first step in this evaluation framework consists in generating the full behaviour of a process model. This entails the generation of all possible variants allowed by the model. In order to make this possible in the presence of loops, we set a maximum number of repetitions. Then, in order to test behavioural generalization, we perform a resampling procedure at the level of variants. The main underlying logic is that, even by leaving out a fraction of the possible variants for training, the LSTM should still be capable of learning the process structure. 


Concretely, we single out all cases belonging to one or a few variants into the \textit{Test Log (Te)}. The remaining traces make up the \textit{Training Log (Tr)}. If we decide to only retain the cases pertaining to one single variant in the Test Log, we call this the ``leave-one-variant-out cross-validation'' or LOVOCV mode of our framework. This mode can be seen as the most lenient setup towards assessing LSTMs' generalization power, given that all but one variant are still available for training, thus an almost complete event log. However, the framework also includes the option the separate out multiple variants into the Test Log, which allows for more stringent yet more realistic levels of event log completeness for training. The LSTM model is trained by using prefixes as input and the activity label of the subsequent next event as target output, as described in Section \ref{preliminaries}. The cases belonging to the \textit{Training Log} are preprocessed into all possible prefixes for training the model, together with the targets corresponding to the next event. These prefixes and targets are subsequently split into the \textit{Training Prefixes} and \textit{Validation Prefixes}, by an 80\%-20\% split. The \textit{Validation Prefixes} are only used for early stopping. Once trained, we use the RNN model to simulate a \textit{Simulated Log (Sim)} as follows. We start by presenting the RNN with an almost empty prefix only containing the beginning of sequence token (BOS), padded to the correct input length. As output, the model returns a multiclass vector, providing a probability for each of the possible activity labels to be the next event. Using these probabilities, we can then sample a possible next event, to be appended to the existing prefix. Subsequently, the newly obtained prefix is used to sample a next event in the same way, and so forth, until we reach the end of sequence token (EOS) or a certain predetermined maximum size is reached. LSTMs have been used as generative models similarly in~\cite{Camargo_2019}. 
 
\subsection{Metrics}
The \textit{Simulated Log} should ideally represent an event log that is behaviourally highly similar to the event log that we started from. That is, we expect that, even when leaving out one single variant, the RNN should be able to (1) generalize this variant from the observed variants in the Training Log, (2) avoid creating variants that were not observed in the original event log (Train+Test), and (3) contain all the variants present in the Training Log. Accordingly and following these criteria, three metrics were designed, representing Process Model Structure Learning Fitness ($F_\textit{PMSL}$), Process Model Structure Learning Precision ($P_\textit{PMSL}$), and Process Model Structure Learning generalization ($G_\textit{PMSL}$). Each of these metrics outputs a value between 0 and 1, with a higher value indicating a better outcome. Beware that in this current setting, metrics only make sense when the original \textit{Train+Test Log} and the \textit{Simulated Log} contain the same amount of traces as they make use of nominal counts.  If not, the metrics will have to be corrected, which could be easily done by a correction term, related to the proportions of the sizes of the different logs. 

\paragraph{Process Model Structure Learning Fitness ($F_\textit{PMSL}$)} 
First of all, the fitness metric measures whether the RNN learns and replicates all of the behavior found in the \textit{Training Log}, by measuring to what extent all of the variants present in the \textit{Training Log} are also present in the \textit{Simulated log}. This becomes:
\begin{equation}
\label{eqn:fitness}
    F_\textit{PMSL} = \sum_{v\in\textit{Var(Tr)}} \frac{\textit{Min}\left(\textit{Occ(}v\textit{,Sim)}\textit{, Occ(}v\textit{,Tr)}\right)}{\abs{\textit{Tr}}}
\end{equation}
with, as mentioned before, $\abs{Tr}$ denoting the number of traces in the training event log $Tr$, $\textit{Var(Tr)}$ denoting the set of variants of $Tr$ and $\textit{Occ(v,Tr)}$ and $\textit{Occ(v,Sim)}$ a function denoting the frequency or multiplicity of a variant $v$ in these event logs. This metric thus also explicitly takes into account the frequencies of the variants in the different logs. For fitness, this means that the multiplicity of each variant in the \textit{Simulated Log} is expected to be equal to the frequency of observation of that variant in the \textit{Training Log}. Therefore, the fitness measure will punish if a certain variant is under-represented in the \textit{Simulated Log}. 

\paragraph{Process Model Structure Learning Precision ($P_\textit{PMSL}$)} Secondly, the precision metric measures whether the RNN allows for too much behaviour, i.e. traces that have not been seen in \textit{Train+Test Log}. Frequency-wise as well, if certain correct variants are over-represented in the \textit{Simulated Log} the precision will also decrease, resulting in: 
\begin{equation}
\label{eqn:precision}
    P_\textit{PMSL} = \sum_{v\in\textit{Var(Sim)}} \frac{\textit{Min}\left(\textit{Occ(}v\textit{,Sim)}\textit{, Occ(}v\textit{,Tr+Te)}\right)}{\abs{\textit{Sim}}}.
\end{equation}

\paragraph{Process Model Structure Learning generalization ($G_\textit{PMSL}$)} 
Finally, the generalization metric quantifies to which extent the RNN is able to generalize, i.e. whether it is able to learn and reproduce correct but unseen behaviour. Therefore, the metric also measures whether the frequency of occurrence of the unseen variant(s) in the \textit{Test Log} is actually reproduced to the same level in the \textit{Simulated Log}. This becomes:
\begin{equation}
\label{eqn:generalization}
    G_\textit{PMSL} = \sum_{v\in\textit{Var(Te)}} \frac{\textit{Min}\left(\text{Occ(}v\textit{,Sim)}\textit{, Occ(}v\textit{,Te)}\right)}{\abs{\textit{Te}}}.
\end{equation}

In this work, we opt to base our metrics on the multiplicities of the variants. This is a deliberate choice. Nonetheless, one might be less interested in the actual frequency of observation of a variant, but rather in the fact whether a variant is present or not. Accordingly, in Appendix \ref{extra}, we introduce three additional frequency-agnostic metrics, which avoid taking the actual multiplicities of each variant into account. While these metrics are a possible alternative, we consider them less robust as they tend to be highly dependent on the size of the \textit{Simulated Log} and the original \textit{Train+Test Log}. Moreover, these frequency-agnostic metrics can be considered more stringent, particularly in terms of fitness and precision, given that missing a variant in the training or introducing a new variant in the test, albeit with a very low frequency, will have a more pronounced negative effect on these metrics.

\section{Experimental Setup}\label{Setup}
To address the four hypotheses introduced in Section \ref{sec1}, this section details our experimental setup. First, we introduce two sets of process models. Next, the hyperparameter configuration of the LSTM models is discussed. Finally, the different experiments are introduced.


\subsection{Process Models}
We use two sets of process models. 

\begin{figure*}[ht]
\captionsetup[subfigure]{labelformat=empty}
\begin{subfigure}{0.49\linewidth}
  \includegraphics[width=\linewidth]{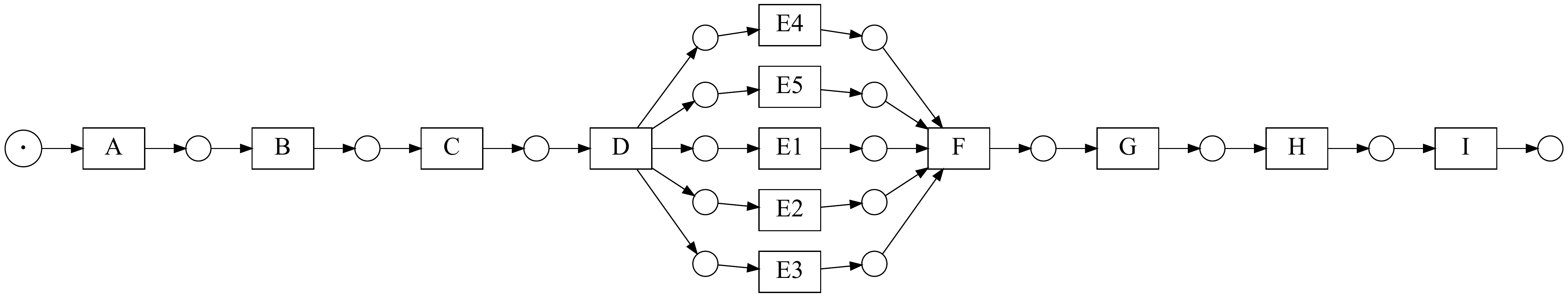}
\caption*{Model 1: Five-branch parallel split}
\end{subfigure}
\hfill
\begin{subfigure}{0.49\linewidth}
  \includegraphics[width=\textwidth]{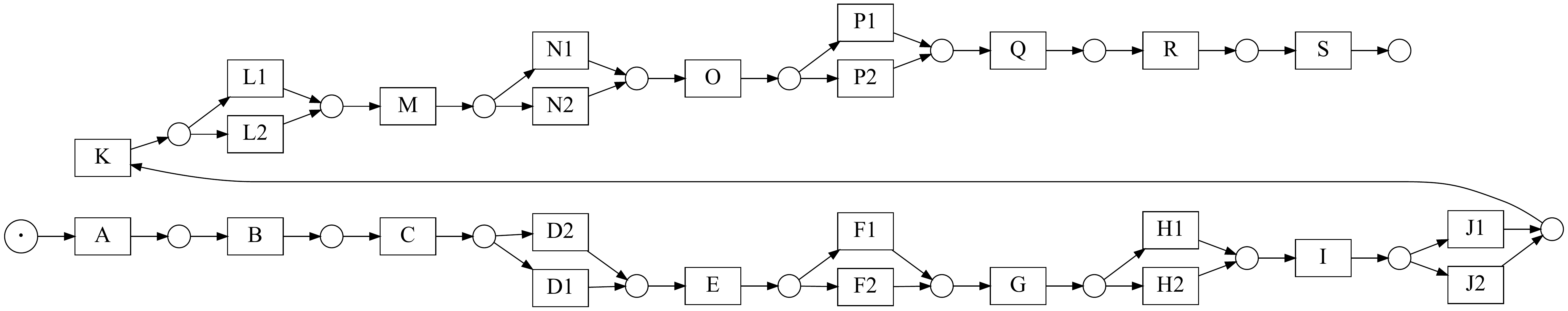}
    \caption{Model 2: Seven consecutive XOR splits}
\end{subfigure}
\hfill
\begin{subfigure}{0.49\linewidth}
  \includegraphics[width=\textwidth]{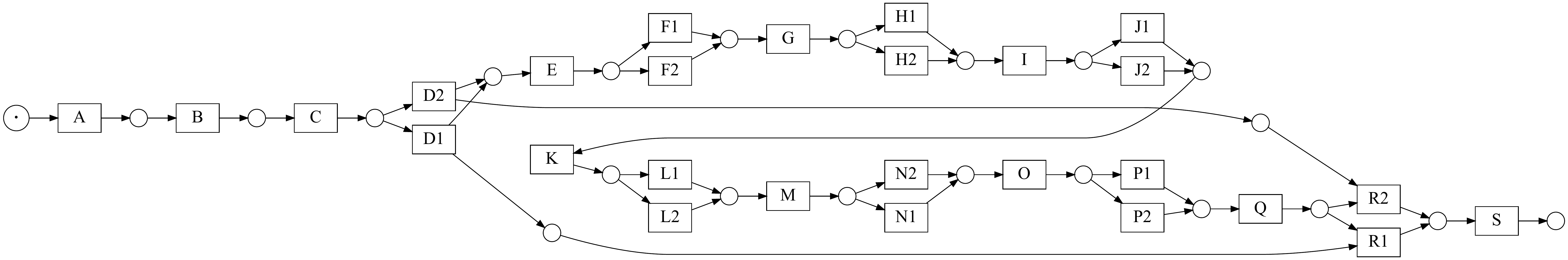}
\caption{Model 3: Eight consecutive XOR splits and long term dependency}
\end{subfigure}
\hfill
\begin{subfigure}{0.49\linewidth}
  \includegraphics[width=\textwidth]{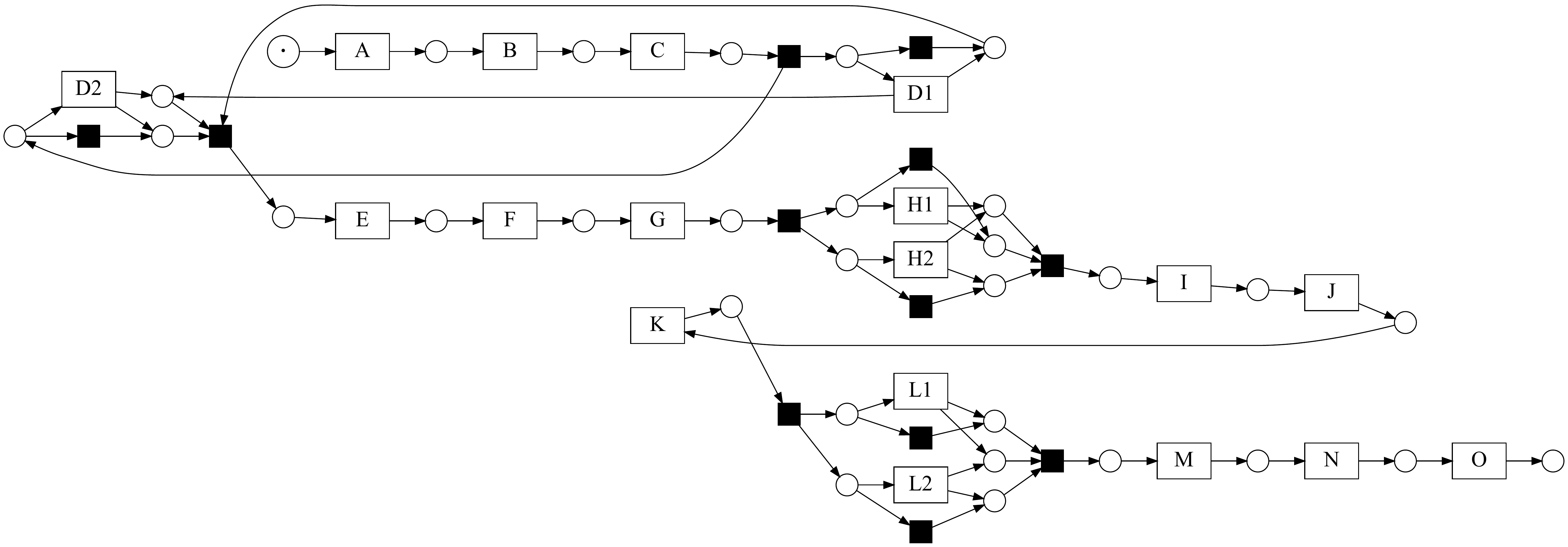}
  \caption{Model 4: Three consecutive IOR splits}
\end{subfigure}
\hfill
\begin{subfigure}{0.49\linewidth}
  \includegraphics[width=\textwidth]{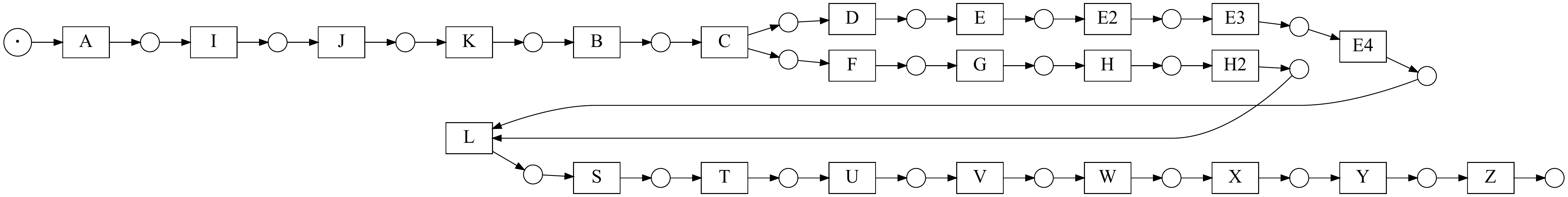}
 \caption{Model 5: Two-branch parallel split}
\end{subfigure}
\hfill
\begin{subfigure}{0.49\linewidth}
\includegraphics[width=\textwidth]{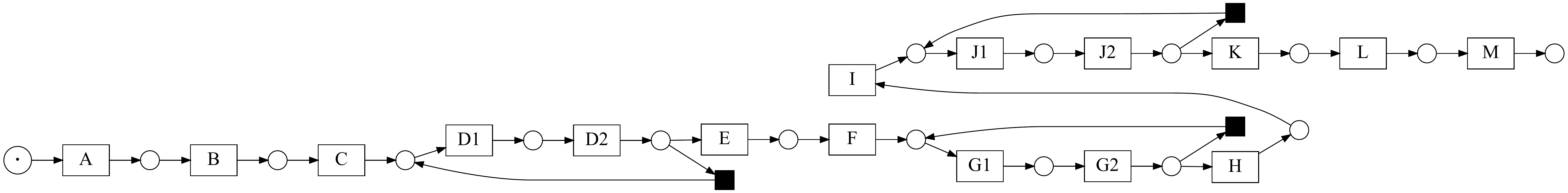}
\caption{Model 6: Three consecutive length-two loops}
\end{subfigure}
\hfill
\vspace{2mm}
 \caption{The six constructed basic \textit{sine qua non} process models}
\label{fig:models}
\vspace{-6mm}
\end{figure*}

\paragraph{Set 1: Six basic \textit{sine qua non} models}
To start with, we have carefully constructed six process models to provide a \textit{sine qua non} condition in terms of investigating whether LSTM models can deal with a number of basic control-flow patterns: parallelism, exclusive choice, long term dependency, inclusive choice, and loops. These deliberately simple models are shown in Figure \ref{fig:models}. Model 1 contains a classical parallel block structure consisting of five branches containing each one single activity, resulting in a process that has 120 (equally likely) control-flow variants. In Models 2 and 3, two process models with 128 (equally likely) control-flow variants are created by sequencing seven and eight exclusive OR (XOR) splits respectively, but with a long-term dependency (LTD) added to the last split in Model 3, connecting it to the first split. Model 4 consists of three inclusive OR (IOR) splits, where at least one, but possibly both of the two activities can occur. This leads to in total 64 control-flow variants. Model 5 shows a different variation of parallel behavior, by introducing parallel block structure that has only two parallel paths, yet consisting of 4 and 5 activities respectively, leading to 126 control-flow variants with varying likelihood. Finally, Model 6 shows a process with three consecutive length-two loops. The number of possible control-flow variants is technically unlimited, however by restricting each marking to be visited a maximum of three times, we reduce this to 27 different variants. An overview of these models can be found in Table \ref{tab:overview} together with the length of the in put prefixes used by the LSTM.

\begin{table}[ht]
\centering
\begin{tabular}{lllllll}
\hline
Model & Control-flow pattern & \# Var & \# Act. & Min. len. & Max. len. & Prefix len.\\ \hline
1    & Parallel    & 120    & 13      & 13        & 13 & 6       \\ 
2    & XOR         & 128    & 26      & 19        & 19 & 6      \\ 
3    & XOR+LTD     & 128    & 27      & 19        & 19 & 18       \\ 
4    & IOR         & 64     & 18      & 15        & 18 & 6       \\ 
5    & Parallel    & 126    & 24      & 24        & 24  & 12      \\ 
6    & Loop        & 27 
& 16      & 16        & 28   & 12     \\ \hline
\end{tabular}
\caption{An overview of the six constructed basic \textit{sine qua non} process models \label{tab:overview}}
\vspace{-6mm}
\end{table}

\paragraph{Set 2: Eleven models with varying degrees of parallelism}
In addition to the set of basic models, a second set of models was designed with a particular focus on varying degrees of parallelism. The primary reason for this second set of models is to allow for testing H\ref{hyp:4}).
These models were selected from a family of process trees generated using different settings, corresponding to different levels of parallelism, of the random Process Tree generation tool as introduced in~\cite{ProcessTree} and implemented in the Python Process Mining library PM4PY~\cite{pm4py}, version 2.2.18. We use multiple different settings corresponding to probabilities for adding different operators each time you add a new node to the tree: keeping the probability to add a sequence operator (SEQ) to 50\%, and each time a different way of distributing the other 50\% over the probabilities of adding a parallel (AND) or choice (XOR) operator respectively. For each setting a plethora of different sized (ranging in between 10 and 25 visible activities) process trees was generated\footnote{In the, since then updated version of the PM4PY~\cite{pm4py} (version 2.2.18, fixed since 2.2.19), silent activities could still be generated in certain cases, even though the setting was put to 0. Because there is no intrinsic reason we do or do not want to consider these, and because certain experiments were already performed, we opted to keep using the models that do contain some silent activities.}. In order to emphasize the difference between parallel and non-parallel behavior, and in this way make the experiment more controllable, it was opted to use the exclusive OR (rather than the inclusive OR). Respectively, for each different setting corresponding to a different level of parallelism, we only kept the process models with the number of different possible control-flow variants to be in between 80 and 160. We do this because we want to be able to investigate the effect of parallelism purely from the intrinsic difference between it and other types of behavior, without the influence of having large differences in number of control-flow variants. This is a choice made with hypothesis \ref{hyp:4} in mind. It is important to note that it does prohibit to research the effect of complexity, measured by number of variants a model can produce. In order to only consider models which produce traces of considerable length, another condition is set on the minimum trace length, i.e. for it being longer than 5 activities. From the process trees not filtered out yet, the tree is selected for which the actual proportions of operators (sequence, choice and parallelism) corresponded best with the settings (probabilities for adding each operator). Of course, the exact location of a certain operator (e.g. a parallel operator at the top of a process tree or at the bottom of one of its branches) can also have a big influence on its complexity. In the ideal case multiple models should therefore be tested for each setting, taking these differences into account as well. Due to the computational demands of the experiments however, this was not possible in this work. The selected set of models can be found online\footnote{\url{https://github.com/jaripeeperkorn/LSTM_Process_Model_Structure}} and an overview can also be found in Table \ref{tab:overview2}. For each model, the size of the play-out \textit{Train + Test Logs} is set to be $100$ times the number of possible variants in the model. The LSTMs trained on all of these models use a prefix length of the maximum trace length, subtracted by one.

\begin{table}[ht]
\centering
\begin{tabular}{llllllll}
\hline
Model & SEQ & AND & XOR & \# Var. & \# Act.& Min. len. & Max. len. \\ \hline
1     & 50\%         & 0\%          & 50\%       & 96          & 19            & 7           & 9           \\ 
2     & 50\%         & 5\%          & 45\%       & 108         & 18            & 6           & 10          \\ 
3     & 50\%         & 10\%         & 40\%       & 120         & 19            & 7           & 9           \\ 
4     & 50\%         & 15\%         & 35\%       & 150         & 18            & 7           & 11          \\ 
5     & 50\%         & 20\%         & 30\%       & 136         & 17            & 8           & 10          \\ 
6     & 50\%         & 25\%         & 25\%       & 108         & 11            & 6           & 8           \\ 
7     & 50\%         & 30\%         & 20\%       & 160         & 18            & 7           & 13          \\ 
8     & 50\%         & 35\%         & 15\%       & 140         & 16            & 8           & 13          \\ 
9     & 50\%         & 40\%         & 10\%       & 120         & 11            & 9           & 10          \\ 
10    & 50\%         & 45\%         & 5\%        & 120         & 11            & 11          & 11          \\ 
11    & 50\%         & 50\%         & 0\%        & 144         & 11            & 11          & 11          \\ \hline
\end{tabular}
    \caption{An overview of the 11 process models with varying degrees of parallelism. \label{tab:overview2}}
\vspace{-6mm}
\end{table}

\subsection{Hyperparameter Search}

\subsubsection{Grid}

The generalization capacity of RNNs strongly depends on tuning its hyperparameters. This is an essential part of the training procedure. An overview of the investigated hyperparameters can be found in Table~\ref{tab:grid}. As mentioned earlier, the usage of either a uni- or bidirectional LSTM layer is the first binary hyperparameter. The number of stacked LSTM layers is varied between one and two, with the layers' hidden dimension size set to 16, 32 or 64 units. The model's weights are optimised making use of the Adam~\cite{Adam} algorithm. The mini-batch size was fixed to contain $128$ prefixes. The learning rate is each time started at $0.005$ and decreased (by a factor two) whenever the accuracy on the validation set has not decreased for over 3 epochs, and the training is stopped (early stopping) when the accuracy has not increased for over 6 epochs (or when a maximum of 600 epochs is reached). The choice of optimizer, learning rate and batch size is not added to the grid and was selected based on literature~\cite{Tax_2017, Camargo_2019}. For clarity, the RNN is trained optimising accuracy in a ``classical'' sense, i.e. whether the activity predicted by the model to be the most probable activity is actually correct. Next to early stopping, there are several measures one can take to try and force a neural network to generalize and avoid overfitting. One of these is dropout~\cite{dropout}, in which, each epoch, a fraction of the nodes is randomly selected to be ignored during training, reducing the likelihood of overfitting to the training data by forcing the model to consider ``alternative paths'' during training. In this work we chose to add dropout to the output of the LSTM layer (note that dropout on the inputs would lead to removing certain steps of a prefix which is undesirable in our setting), and experiment with multiple dropout values (including no dropout). We also experiment with \textit{L1} (\textit{Lasso}) and \textit{L2} (\textit{Ridge}) regularisation~\cite{regularization}, which add a small penalty to the model based on the absolute or squared value of the weights respectively. In this paper we use both, but keep them at the same value in order to limit the size of the hyperparameter grid. This led to 192 different hyperparameter configurations to iterate over. The Python library Keras \footnote{\url{https://keras.io}} has been used to implement the LSTMs in this work. We omitted using batch normalisation due to its limited effectiveness when applied to RNNs~\cite{Cooijmans_2016}. In future work, however, exploring with the effectiveness of \textit{Recurrent Batch Normalisation}~\cite{Cooijmans_2016} and \textit{Layer Normalisation}~\cite{Ba_2016} might increase the quality of the LSTMs. The array of parameter values in the grid were selected to keep a balance between testing enough different settings and limiting the (exponentially) growing size of the experiments.

\begin{table}[ht]
    \centering
    \begin{tabular}{c|c}
        \textbf{Hyperparameter} & \textbf{Values} \\ \hline
        Uni- or bidirectional? & One, Two \\
        Number of LSTM Layers (if two directional technically double) & 1, 2 \\
        LSTM Layer Size & 16, 32, 64 \\
        L1 and L2 & 0.0, 0.0001, 0.001, 0.01 \\
        Dropout & 0.0, 0.2, 0.4, 0.6
    \end{tabular}
    \caption{The hyperparameter values used in the grid search. \label{tab:grid}}
\vspace{-6mm}
\end{table}

\subsubsection{Accuracy Optimized Hyperparameters}
For the first hyperparameter configuration, we select optimal settings by means of the default optimization strategy, i.e. finding those hyperparameters that maximize the next-event prediction accuracy (and categorical cross-entropy loss) as tested on a hold-out validation set. Please observe that this validation set is not to be confused with the \textit{Test Log} described in the Section \ref{Metrics}, as there is no variant-based sampling applied, yet the prefixes are split $80-20\%$ randomly. For each process model, this experiment was carried out using the full \textit{Train+Test Log} without variants missing. From the results of this search, it could be noticed that higher regularization and dropout, lead to (slightly) lower accuracy on average. Results over different hyperparameter settings were also often comparable. Since these different accuracy values are similar, it was chosen to pick one common hyperparameter setting for all process models. This was also incentivised because we want to be able to compare the result on different process models properly, without having the effect of (random) differences due to variations in the selected hyperparameters, only slightly better in this particular run. We chose every hyperparameter such that the average loss (over all models trained with this parameter) was minimal. \textbf{This resulted in a bidirectional, one layered LSTM with size $64$ and no regularization and no dropout}. Note that the good scores for this setting is likely to be a result of overfitting. To be clear, this particular combination of all parameters does not (necessarily) lead to the lowest loss out of all possible setting, nor highest accuracy, for all process models. But for each parameter individually, it was checked that the choices did result in the lowest (average) loss (over all models trained with that parameter put to that particular setting), and this both on average, as well as for each of the process models individually. \\

\subsubsection{Post-hoc Optimized Hyperparameters}
A second hyperparameter configuration consists in a post-hoc optimization in function of the metrics defined in this work. Please observe that this optimization is not possible in practice, yet in order to test the best possible performance of LSTM models, it is a worthwhile approach in this work, in line with our preliminary work in~\cite{Jari}. In order to contain computational requirements, only a grid search applied to Model 1 of Figure \ref{fig:models} was carried out, with the best hyperparameter settings for Model 1 being subsequently applied to all other process models. 


We conducted a tailored leave-one-variant-out cross-validation (LOVOCV) procedure as introduced above. In each LOVOCV iteration, we singled out all cases pertaining to one single variant into the \textit{Test Log}. For every hyperparameter combination, we performed the tailored LOVOCV eight times, each time with a different variant in the test set. In each iteration, we obtained a \textit{Simulated log} of equal size of the original \textit{Train+Test Log}. Based on these eight LOVOCV iterations, we calculated the three different metrics defined above, and, for each setting, took the average over the eight iterations. Note that we use the \textit{Test Log} for this hyperparameter tuning, rather than the \textit{(cross) validation log} as is usual. Since we are not trying to compare the predictive quality of different approaches as such, this is justified. We choose to continue working with the setting showing the highest average score over all three metrics, \textbf{i.e., bidirectional, one LSTM layer of hidden size 64, an \textit{L1} and \textit{L2} of $0.001$ and a \textit{dropout} of $0.4$}. Because this model was only trained on the data of one simple process model, and the differences between certain settings were slim, we however do not want to claim this setting is ideal for each predictive process monitoring problem. However in the context of this comparative investigation, which does not try to claim presenting a more optimal approach, we deem it permitted to continue with this setting for all models in this paper.

\subsection{Experiments}

For all of the experiments we work with a \textit{Train+Test Log} with a size of 100 times the number of play-out variants the model can produce. We first start by testing the most lenient setting, i.e. whether the LSTM models are able to reproduce one variant when this variant is left out of the \textit{Training Log}. This is the so-called ``leave-one-variant-out cross-validation'' or LOVOCV. This LOVOCV setting is performed on each of the process models introduced earlier. As for some models the frequency of the variants is not evenly distributed and to obtain more robust results, we  conducted an exhaustive LOVOCV, i.e. the procedure is repeated as many times as the number of variants in the event log, so that every variant is used once to form the \textit{Test Log}. We do this exhaustive LOVOCV for both hyperparameter setting described above, from which we might provide us with the possibility to confirm both hypothesis \ref{hyp:1} and \ref{hyp:2} since the different settings display different levels of overfitting countermeasures. By performing these experiments for the 11 generated process models, each with a different level of parallelism, we might also be able to gain insights concerning hypothesis \ref{hyp:4}. 

Furthermore, we have also performed some additional experiments in which different amounts of variants are set aside in the \textit{Test Log}. We have done these in an exhaustive manner as well, meaning that we each time design different folds to be used as \textit{Test Log} in such a way that every variant has been put into the \textit{Test Log} exactly once. The LOVOCV setting can be regarded as having the amount of folds equal to the number of variants. Furthermore the different amount of folds used are chosen to be: 20, 15, 10, 8, 6, 5, 4, 3 and 2. To limit the scope of our research paper topic, these experiments are intentionally only shown as performed on the six (simple) models from Table \ref{tab:overview}. However, since they display different types of model complexity, this will still allow us to investigate hypothesis \ref{hyp:3}. 

\section{Results and Evaluation}\label{Results}
In this Section, we show the results of the different experiments described above. We first discuss the results of the LOVOCV experiments, starting with the first family of models from Table \ref{tab:overview}, followed by the generated ones from \ref{tab:overview2}, each time for both aforementioned hyperparameter settings. The metric scores produced by the LSTMs trained with these two settings are compared in Table~\ref{tab:Results1} and \ref{tab:Results2}. Subsequently the experiments with the different-sized folds are shown. 

\subsection{Leave-One-Variant-Out Cross Validation (LOVOCV)}

\subsubsection{Set 1: Six basic \emph{sine qua non} models}

Performing the LOVOCV experiment allows us to inspect the LSTMs' learning capability in the most lenient setting, since we are working with \textit{Test Logs} containing only a single variant. Moreover, each of the simple models in Table~\ref{tab:overview} is restricted to a single type of control-flow behavior. Therefore, we can relate the learning capability to the type of behavior. 

Table \ref{tab:Results1} shows the results for the exhaustive LOVOCV applied to these models. In the case of Model 6 (loops), we restricted the analysis to variants for which each loop is taken a maximum of three times (27 variants). For each metric, the average values over all folds are displayed together the standard deviation. On the left, the results are displayed for the LSTMs trained with the aforementioned next event prediction accuracy based hyperparameters, while on the right the scores for the LSTMs trained with the post-hoc hyperparameter setting can be seen. The difference between the two (especially for generalization scores) is striking. Several interesting observations can already be made from these results on the left. First, models displaying parallel behavior (Models 1, 5 and to a lesser extent 4) or a long-term dependency (Model 3) seem to be problematic for the LSTMs in terms of generalization. Since, in contrast, fitness ($F_{PMSL}$) and precision ($P_{PMSL}$) scores are high, the most likely explanation is that the LSTMs are simply overfitting, only memorizing what they have previously seen during training without learning any process structure. For the other models (Model 2 and Model 6), corresponding  to XOR and looping behavior respectively, better generalization is achieved. However, given the (extremely) lenient setting of the LOVOCV, it is still remarkable that these scores do not go well above $0.8$. Combined with the results on the less robust metrics in Appendix \ref{extra}, which simply measure whether certain variants are detected at all, we remark that the lower generalization scores of Model 2, 4 and 6 are due to the \textit{Simulated Log} not containing the left out test variant \textit{enough}, i.e. less then in the Petri Net play-out, rather than not at all. Accordingly, intrinsically the LSTM is able to learn behavior not seen in training, however the frequency of occurrence is well below of what is expected. Overall these results show that LSTMs built for next event prediction and optimized for prediction accuracy, do not properly learn process structure. This is most obvious for models containing parallel constructs. even despite the utterly simple and stylized models. 

\begin{table}[ht]
    \centering
    \setlength\tabcolsep{3pt}
    \resizebox{\textwidth}{!}{%
    \begin{tabular}{ccc|ccc|ccc}
        \hline
        &&& \multicolumn{3}{c|}{Acc. based Hyperparameters}
        & \multicolumn{3}{c}{Post-Hoc Hyperparameters}\\    
        \textbf{Mod.}  & \textbf{Pattern} & \textbf{\#Var.} & \textbf{$F_\textit{PMSL}$} & \textbf{$P_\textit{PMSL}$}  &\textbf{$G_\textit{PMSL}$} & \textbf{$F_\textit{PMSL}$} & \textbf{$P_\textit{PMSL}$}  &\textbf{$G_\textit{PMSL}$} \\ \hline
        1 & PAR & 120 & $0.96\pm0.01$ & $0.95\pm0.01$ & $0.00\pm0.01$ & 
        $0.94\pm0.00$ & $0.94\pm0.00$ & $0.84\pm0.11$  \\ 
        2 & XOR& 128 &$0.94\pm0.01$ & $0.94\pm0.01$ & $0.80\pm0.14$ & 
        $0.94\pm0.00$ & $0.94\pm0.00$ & $0.92\pm0.09$  \\ 
        3 & XOR+LTD & 128 &$0.95\pm0.01$ & $0.94\pm0.00$ & $0.04\pm0.17$& 
        $0.94\pm0.00$ & $0.94\pm0.00$ & $0.92\pm0.10$  \\ 
        4 & IOR & 64 &$0.93\pm0.02$ & $0.92\pm0.02$ & $0.66\pm0.17$& 
        $0.92\pm0.01$ & $0.92\pm0.01$ & $0.71\pm0.14$   \\ 
        5 & PAR & 126 &$0.95\pm0.02$ & $0.94\pm0.02$ & $0.07\pm0.20$ & 
        $0.90\pm0.01$ & $0.90\pm0.01$ & $0.67\pm0.21$  \\ 
       6 & LOOP & 27 &$0.87\pm0.02$ & $0.86\pm0.02$ & $0.75\pm0.28$ & 
        $0.86\pm0.02$ & $0.85\pm0.02$ & $0.86\pm0.21$  \\ \hline
    \end{tabular}}
    \caption{The results on the six \emph{sine qua non} process models from Table \ref{tab:overview}, averaged over all leave-one-variant-out experiments with every different control flow variant. \label{tab:Results1}}
\vspace{-6mm}
\end{table}

On the right of Table \ref{tab:Results1} the scores can be seen for the LSTMs, which have been trained using the hyperparameter settings which were obtained by a post-hoc selection on LSTMs trained on Model 1. These models seems to generalize remarkably better than their accuracy optimized counterparts, scoring significantly better on $G_{PMSL}$. This can be attributed to the explicit overfitting countermeasures ($L1$, $L2$ and $dropout$). The negative impact of these measures on $F_{PMSL}$ and $P_{PMSL}$ also seems to be rather limited. Particularly for the parallel models (Model 1 and 5) and the model showing a long-term dependency (model 3) a major improvement in generalization can be observed. By comparing with the $G_{A-PMSL}$ to the $G_{A-PMSL}$ scores in Table \ref{tab:Results3} (in Appendix \ref{extra}), it can also be concluded that the LSTMs are intrinsically capable of learning every type of the listed simple control-flow behavior. Albeit with lower probabilities for unseen examples, most notably when the process model displays parallel like behavior. Overall, the difference in generalization between the LSTMs, trained with different hyperparameters, seem to support the statement of hypothesis \ref{hyp:2}. 

\paragraph{Timing}

Since the contribution of this work is not to produce a new or better model, but rather understand how existing models learn patterns from data, we do not perform a full efficiency and timing experiment. However, purely indicative we show the efficiency of training an LSTM with both the accuracy based and post-hoc hyperparameter settings on \textit{Model 3}. This is the model where we use the maximum trace length as prefix length and therefore for which the training takes the longest. These timing experiments are performed on an \textit{Intel(R) Xeon(R) CPU @ 2.00GHz}. During the full experiments different GPUs were also used to speed up the training. For the timing of the accuracy based hyperparameter setting training continued for 18 epochs with a training time of 3min 6s. For the post-hoc hyperparameters training continued for a longer period, namely a 100 epochs, with a total training time of 17min 58s. In both cases the simulation of the simulated log took around 1 minute.  

\subsubsection{Set 2: Eleven process models with varying degrees of parallelism}

Performing the same lenient exhaustive LOVOCV described above, on the increasingly more parallel models described in Table \ref{tab:overview2}, could allow us to test hypothesis \ref{hyp:4} to a certain extent, while also confirming what we already found concerning hypothesis \ref{hyp:1} and \ref{hyp:2}. Looking to the left of Table \ref{tab:Results2}, where the metrics as calculated for the LSTMs, trained with the accuracy based hyperparameters, are displayed, the low generalization scores immediately stand out. Together with the absolute metric scores in Table \ref{tab:Results4} it can be concluded that the behavior displayed in these (more complex) randomly generated processes is not always understood by the LSTM. For some \textit{Test Logs} (variants) total overfitting is displayed and for others the LSTM does generalize to the unseen variant, albeit with significantly lower probability (than in the corresponding play-out). This again seems to confirm hypothesis \ref{hyp:1}. On the right of Table \ref{tab:Results2} the scores obtained with the LSTMs, trained with the settings corresponding to the (simplified) post-hoc optimization, are displayed. Again these scores seem to confirm hypothesis \ref{hyp:2} by showing significantly better generalization scores (as compared to their counterparts on the left), without compromising fitness and precision. Together with Table \ref{tab:Results4} we can also conclude that the non-perfect generalization scores are (almost) always due to non-equal simulation probabilities of the unseen variant (as compared to original play-out), rather than not being simulated at all. However, there seems to be no correlation present between the generalization scores and the level of parallelism displayed in these models. We can therefore not confirm hypothesis \ref{hyp:4}. Nonetheless, the results in Table \ref{tab:Results1} do show some differences between the different types of process structure complexity. It is still likely that the proficiency of an LSTM to learn a process model is related to the complexity of the process structure. A more extensive experiment, with possibly no or less stringent conditions on the number of variants, might unveil some ambiguity. This experiment might still related to parallelism to certain extend, since there is some relation between parallelism and number of variants produced by a process model, which most likely can not be ignored.

\begin{table}[ht]
    \centering
    \setlength\tabcolsep{3pt}
    \begin{tabular}{cc|ccc|ccc}
        \hline
        & & \multicolumn{3}{c|}{Acc. based Hyperparameters}
        & \multicolumn{3}{c}{Post-Hoc Hyperparameters}\\  
        \textbf{Par.} & \textbf{\#Var.} & \textbf{$F_\textit{PMSL}$} & \textbf{$P_\textit{PMSL}$}  &\textbf{$G_\textit{PMSL}$} & \textbf{$F_\textit{PMSL}$} & \textbf{$P_\textit{PMSL}$}  &\textbf{$G_\textit{PMSL}$} \\ \hline
        0\% & 96 
        & $0.95\pm0.01$ & $0.94\pm0.01$ & $0.08\pm0.16$ 
        & $0.94\pm0.01$ & $0.94\pm0.01$ & $0.79\pm0.12$ \\ 
        5\% &108
        & $0.95\pm0.01$ & $0.95\pm0.01$ & $0.17\pm0.25$ 
        & $0.95\pm0.01$ & $0.94\pm0.01$ & $0.68\pm0.22$ \\ 
        10\% &120
        & $0.95\pm0.01$ & $0.94\pm0.01$ & $0.13\pm0.24$ 
        & $0.94\pm0.00$ & $0.94\pm0.00$ & $0.78\pm0.14$ \\ 
        15\% &150
        & $0.95\pm0.01$ & $0.95\pm0.01$ & $0.34\pm0.39$ 
        & $0.95\pm0.00$ & $0.94\pm0.01$ & $0.76\pm0.19$ \\ 
        20\% &136
        & $0.96\pm0.01$ & $0.95\pm0.01$ & $0.28\pm0.35$ 
        & $0.95\pm0.00$ & $0.95\pm0.01$ & $0.74\pm0.24$ \\ 
        25\% &108
        & $0.95\pm0.02$ & $0.94\pm0.02$ & $0.16\pm0.29$ 
        & $0.93\pm0.01$ & $0.93\pm0.01$ & $0.58\pm0.25$ \\ 
        30\% &160
        & $0.95\pm0.01$ & $0.94\pm0.01$ & $0.15\pm0.25$ 
        & $0.94\pm0.00$ & $0.94\pm0.00$ & $0.75\pm0.14$ \\ 
        35\% &140
        & $0.95\pm0.01$ & $0.94\pm0.01$ & $0.28\pm0.36$ 
        & $0.94\pm0.00$ & $0.94\pm0.01$ & $0.75\pm0.18$ \\ 
        40\% &120
        & $0.95\pm0.01$ & $0.95\pm0.01$ & $0.07\pm0.14$ 
        & $0.94\pm0.00$ & $0.94\pm0.01$ & $0.75\pm0.15$ \\ 
        45\% &120
        & $0.95\pm0.01$ & $0.94\pm0.01$ & $0.07\pm0.11$ 
        & $0.94\pm0.01$ & $0.94\pm0.01$ & $0.79\pm0.13$ \\ 
        50\% &144
        & $0.95\pm0.02$ & $0.94\pm0.02$ & $0.22\pm0.33$ 
        & $0.93\pm0.01$ & $0.93\pm0.01$ & $0.72\pm0.21$ \\ \hline
    \end{tabular}
    \caption{The results on the different process models from Table \ref{tab:overview2}, averaged over all leave-one-variant-out experiments with every different control flow variant. \label{tab:Results2}}
\vspace{-6mm}
\end{table}

\subsection{Leave-Multiple-Variants-Out Cross Validation}

\begin{figure*}[!htb]
\captionsetup[subfigure]{labelformat=empty}
\begin{subfigure}{0.5\linewidth}
  \includegraphics[width=\linewidth]{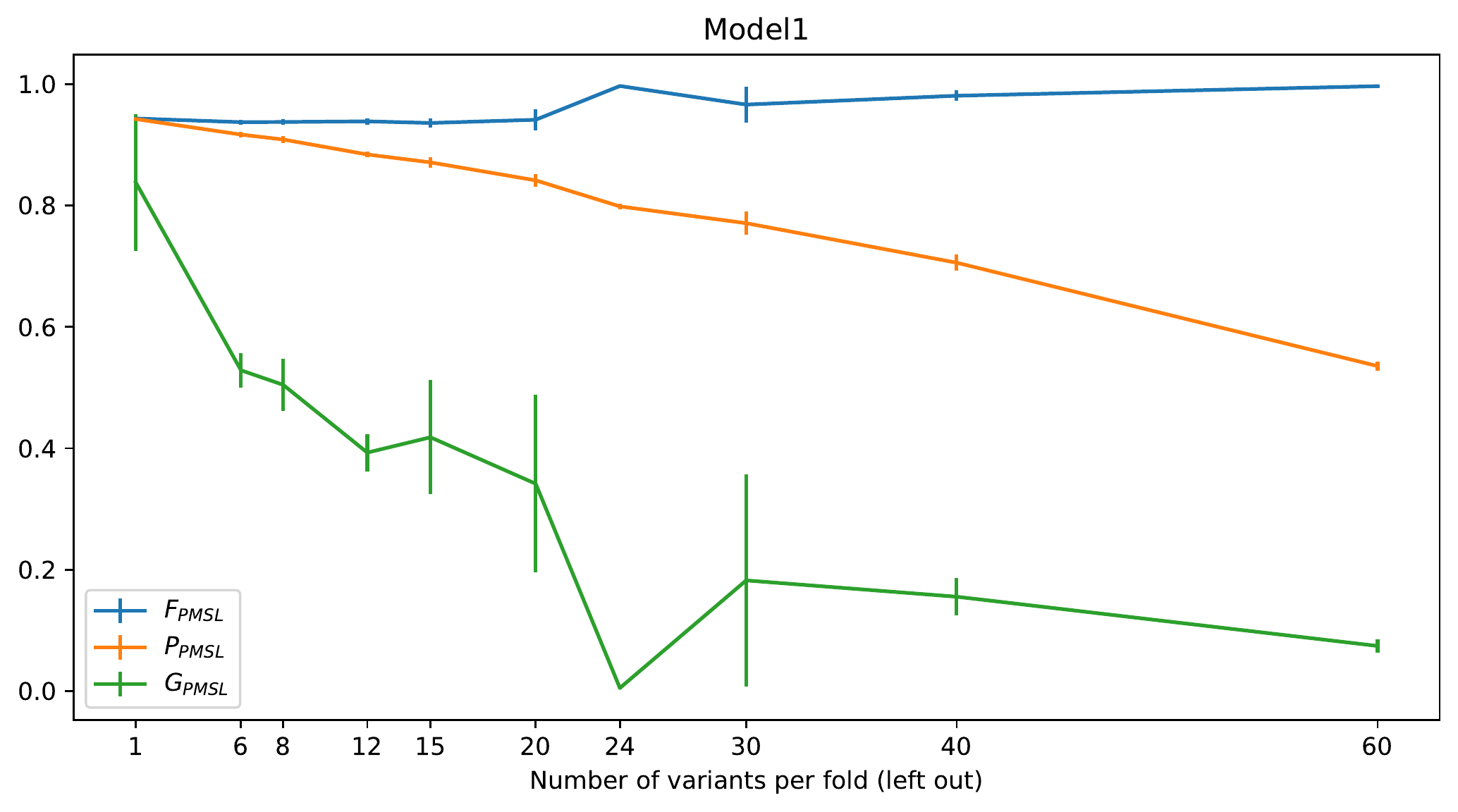}
\end{subfigure}
\hfill
\begin{subfigure}{0.5\linewidth}
  \includegraphics[width=\textwidth]{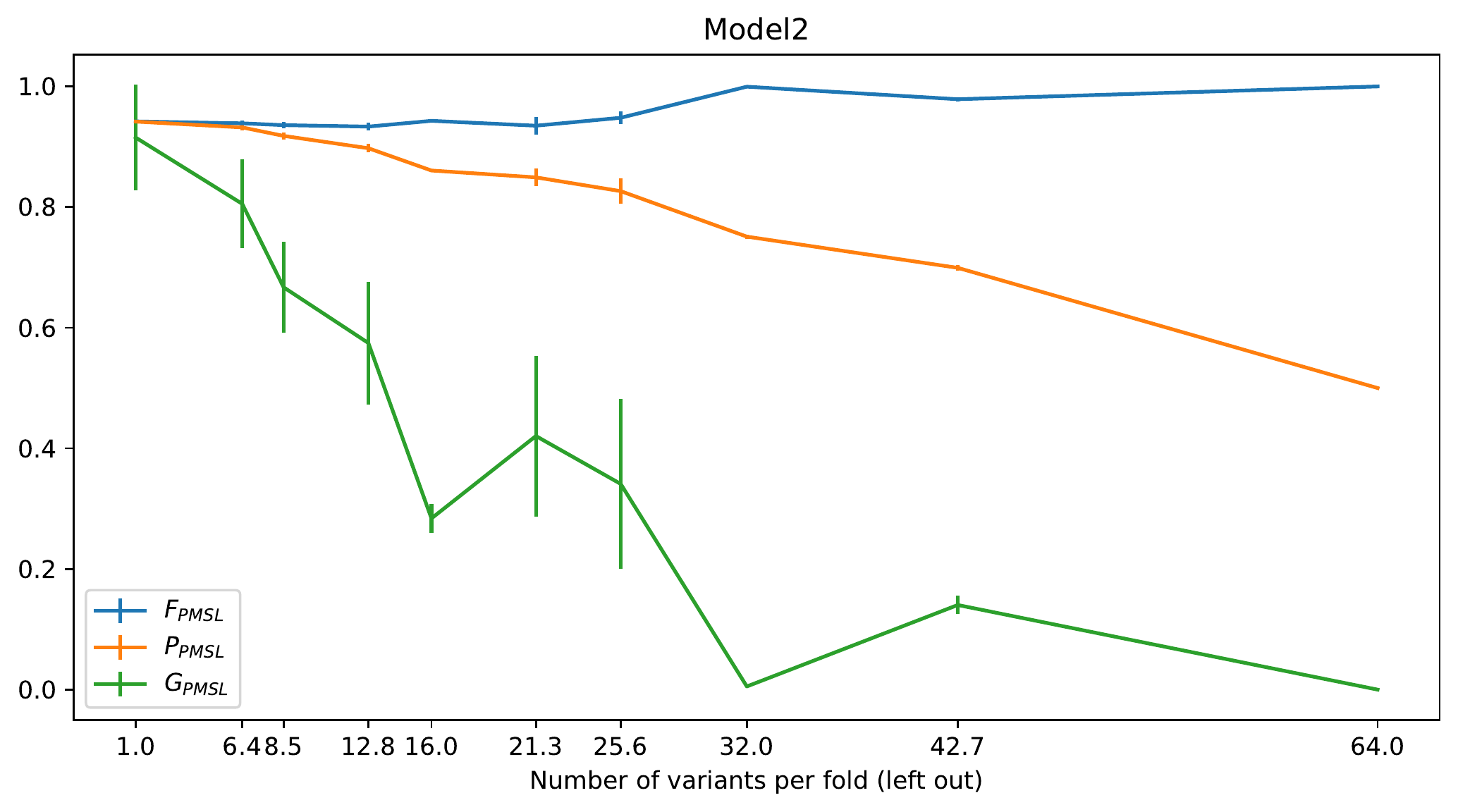}
\end{subfigure}
\hfill
\begin{subfigure}{0.5\linewidth}
  \includegraphics[width=\textwidth]{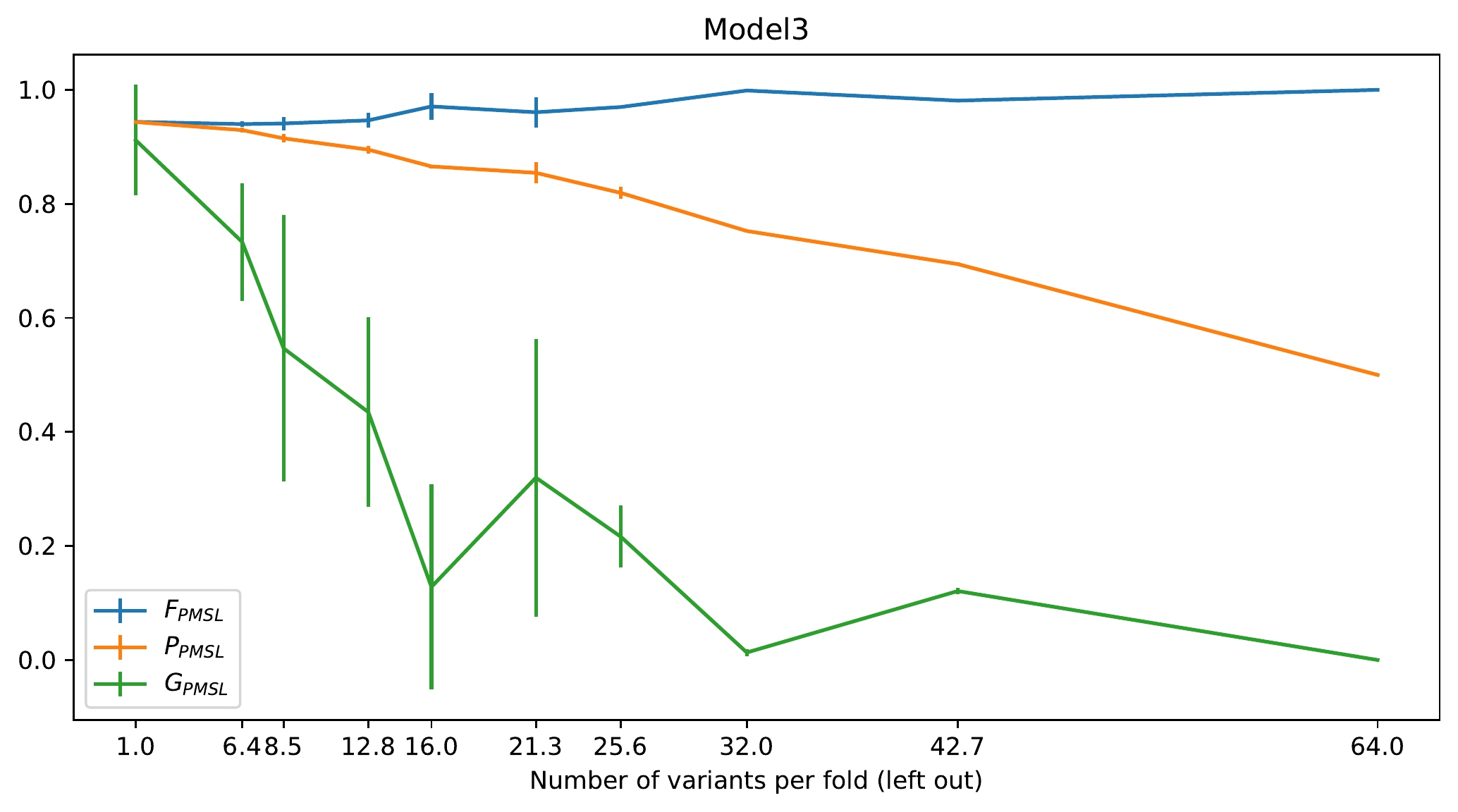}
\end{subfigure}
\hfill
\begin{subfigure}{0.5\linewidth}
  \includegraphics[width=\textwidth]{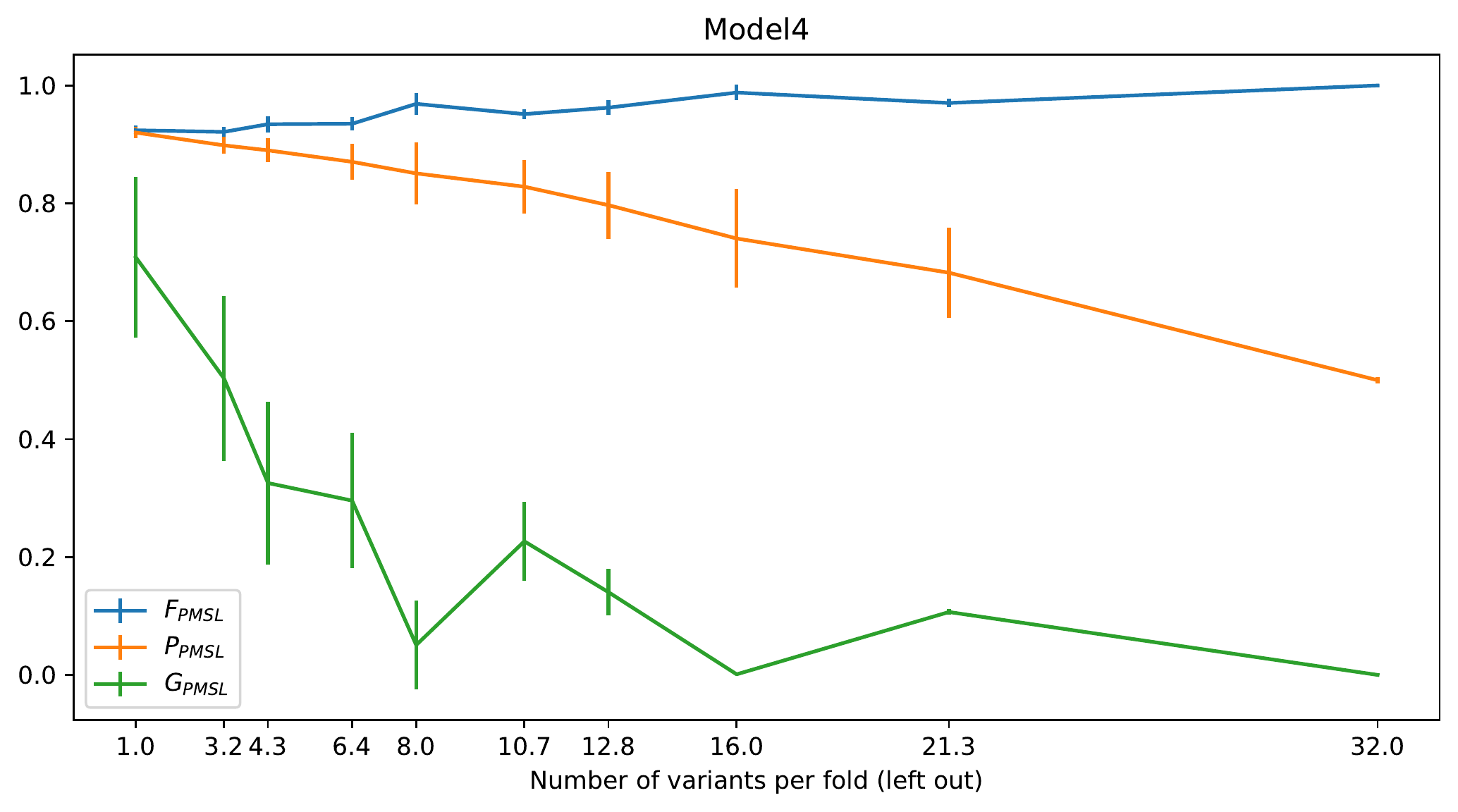}
\end{subfigure}
\hfill
\begin{subfigure}{0.5\linewidth}
  \includegraphics[width=\textwidth]{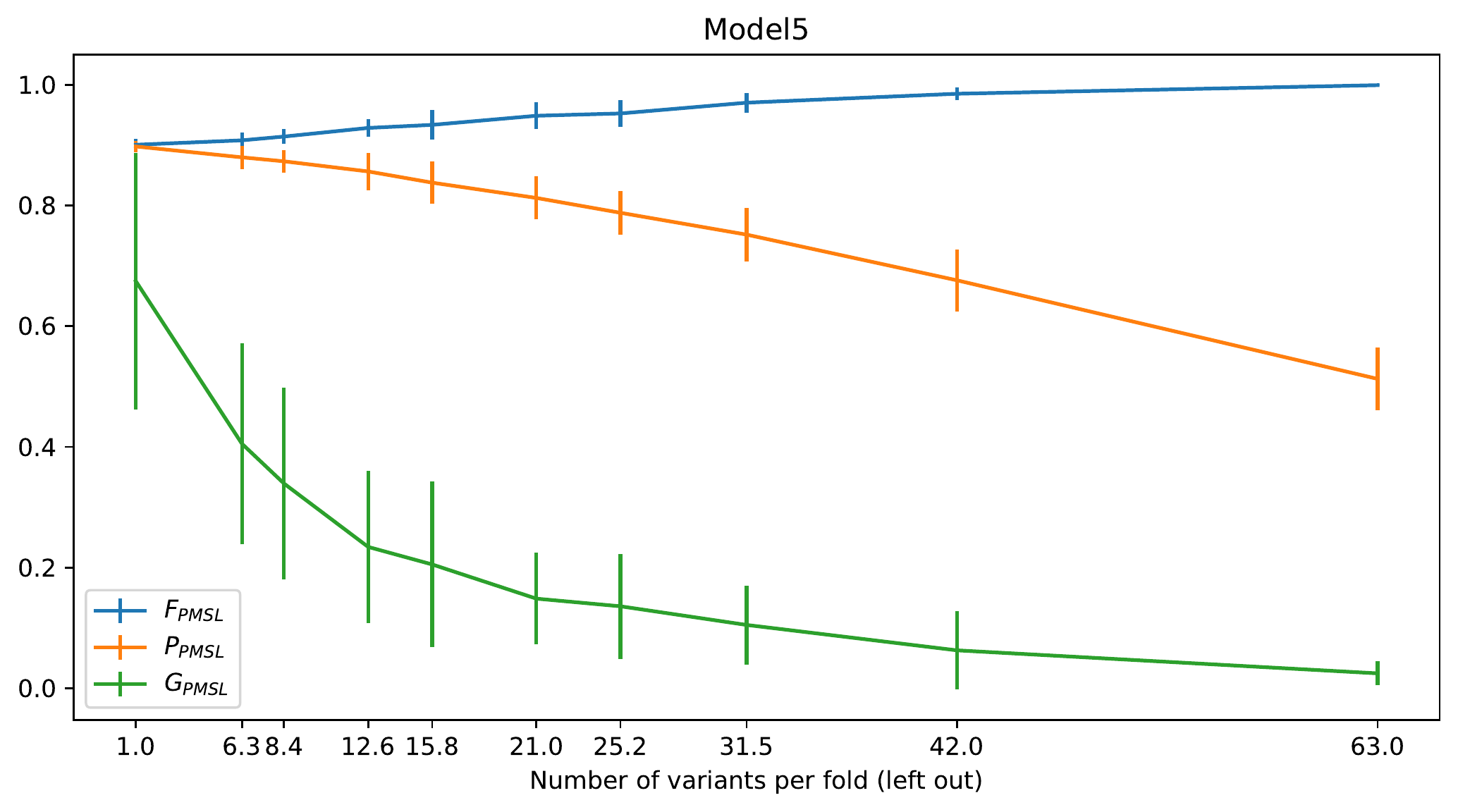}
\end{subfigure}
\hfill
\begin{subfigure}{0.5\linewidth}
\includegraphics[width=\textwidth]{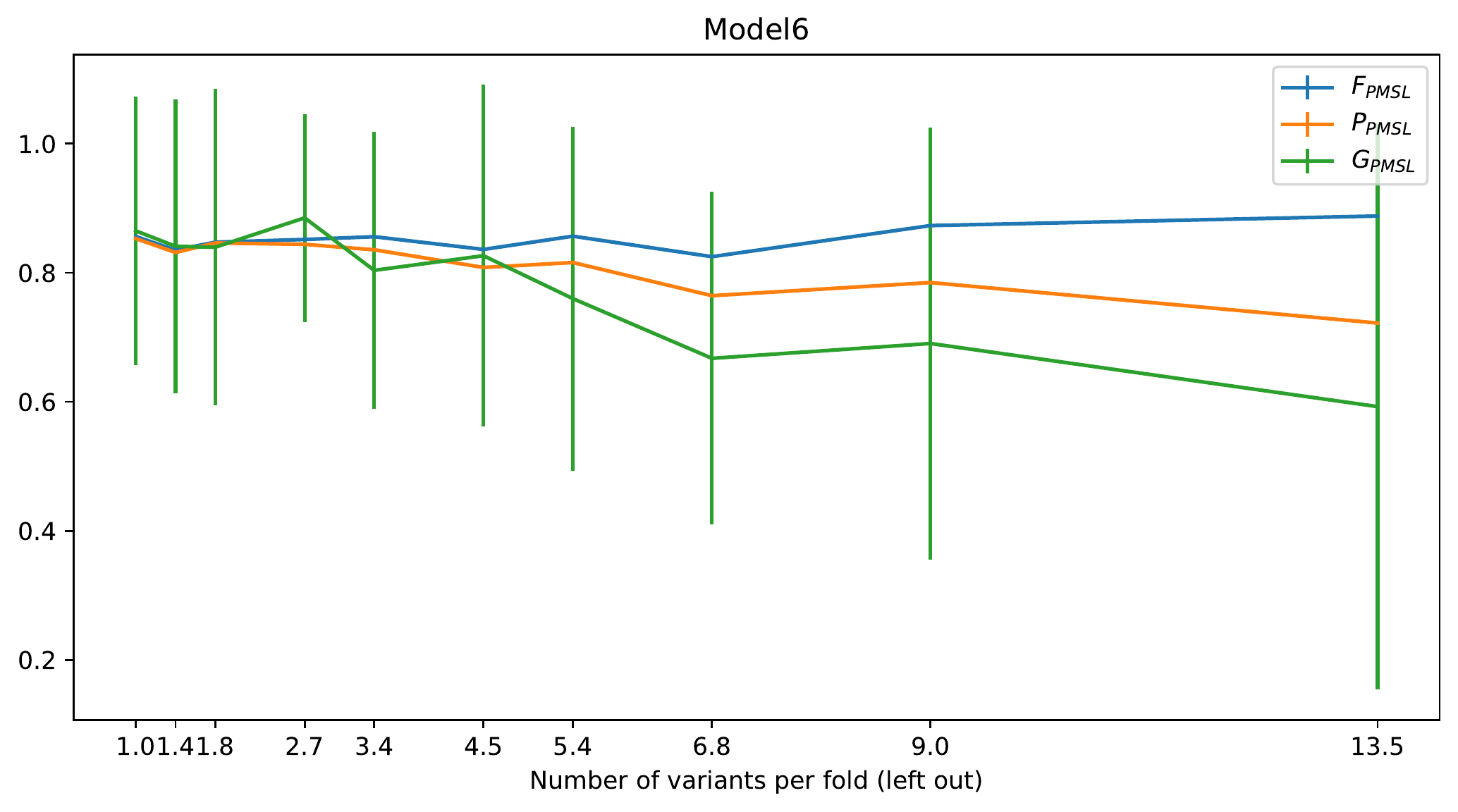}
\end{subfigure}
\hfill
\vspace{2mm}
 \caption{The framework output as performed on LSTMs trained with different amounts of variants in the \textit{Test Log}.}
\label{fig:plots}
\vspace{-6mm}
\end{figure*}

As described earlier, an additional experiment has been performed, concerning \textit{Test Logs} of incrementally increasing sizes. This was achieved by dividing the variants into fewer different folds, in an exhaustive manner (i.e. each variant is put into one fold/\textit{Test Log}). The exact amount of variants per fold depends on the total amount of control-flow variants a process model can produce as well. In Figure \ref{fig:plots} the average metric values (each time over all folds) in function of the amount of variants in these folds is plotted. If the number of variants is not exactly divisible by the number of folds, each variant in the remainder is supplemented to different (randomly selected) folds. Due to the already poor generalization performances of (some of) the LSTMs with accuracy optimized hyperparameters, we opted to train all the LSTMs in this experiment with the so-called post-hoc optimized hyperparameters. 

Figure \ref{fig:plots} shows the results. 
Increasing the amount of variants put aside in the \textit{Test Log}, corresponding to higher levels of incompleteness in the \textit{Training Log}, clearly has an adverse effect on the $G_{PMSL}$ scores, even though LOVOCV results on these models were decent. This confirms hypothesis \ref{hyp:3}. The high standard error values for Model 6 are due to the experiments being significantly more dependent on which exact variant is left out (due to the skewness of the distribution of occurrences of each variant). Precision scores also decrease when leaving out increasing amounts of behavior from the \textit{Training Log}, showing that the decrease in $G_{PMSL}$ is not only due to overfitting. This indicates that the LSTMs models are less capable of learning to understand the process model structure with decreasing information it has to learn from. More precisely, for each of the different types of introduced behaviors, we can see a sharp decrease in $G_{PMSL}$, a slight decrease in $P_{PMSL}$ and a very small increase in $F_{PMSL}$, when increasing the amount of variants left out of he \textit{Training Log}. This decrease in correctly interpreting behavior might be unsurprising, but nonetheless important information considering the sometimes high fractions of incompleteness in real-life event logs~\cite{reallife}. From the same experiment using the alternative metrics in Figure \ref{fig:plotsabsolute}, we can also observe that the decrease in generalization power is not only related to lower simulation probabilities, but as well due to not learning some behavior at all. Next to this, from the constant score of the $F_{A-PMSL}$, the very slight increase in $F_{PMSL}$ can be explained by the fact that the size of the \textit{Simulated Log} stays constant, while overfitting on the behavior in a smaller and smaller \textit{Training Log}. In Appendix~\ref{extra2}, the results are compared to the ability of a more ``classical'' discovery algorithm from the process mining literature to model such types of behavior, by using Process Trees discovered with the Inductive Miner [32]. It is interesting to notice that the ``short'' parallel split in Model 1, causes significantly less issues for the Inductive Miner, compared to how the LSTMs score (``short'' as in the different parallel tracks are only one activity long). However behavior like e.g. the long-term dependency in Model 3 or the inclusive OR split in Model 4, is in general interpreted better by the LSTM models. The results and discussion of this comparative experiment can be found in Appendix~\ref{extra2}.

\section{Conclusion and Future Work}\label{Conclusion}

This paper addressed LSTM neural networks' capability to learn process model structure. Based on a newly developed evaluation framework that combines a variant-level resampling scheme with tailored fitness, precision and generalization metrics, we tested four hypotheses.  

\addtocounter{hyp}{-4}

\begin{hyp}[H\ref{hyp:1}] 
LSTMs built for next event prediction do not properly learn process model structure.
\end{hyp}
\begin{hyp}[H\ref{hyp:2}] 
The incorporation of particular anti-overfitting measures can allow an LSTM to learn process model structure.
\end{hyp}
\begin{hyp}[H\ref{hyp:3}] 
The higher the degree of incompleteness of the training set in terms of control-variants observed, the worse an LSTM will generalize the control-flow behavior.
\end{hyp}
\begin{hyp}[H\ref{hyp:4}] 
The more parallel behavior in a process, the worse an LSTMs’ generalization capability.
\end{hyp}

Regarding H\ref{hyp:1}, we confirmed that even with simplistic models and in a very lenient LOVOCV setup, LSTMs struggle to learn process model structure, as evidenced by the fact that unseen behavior cannot be correctly reproduced. Nonetheless, our results also show that, in terms of H\ref{hyp:2}, anti-overfitting measures can mitigate the problem, however the post-hoc hyperparameter optimization is not realistic for application in practice. As you will need to set your hyperparameters upfront, and not when you confront the model with the future (and possibly unseen) data. As such, finding optimal hyperparameters that can guarantee process model structure learning remains an open problem. Furthermore, H\ref{hyp:3} was confirmed as well, given that increasing the amount of variants (behavior) not seen by the LSTM during training causes the scores for both generalization as well as precision to decrease sharply. This finding has important practical repercussions, given that it is well-known that real-life event logs hardly ever contain complete behavior~\cite{reallife}. Finally, out of the four hypotheses, only H\ref{hyp:4} was rejected. In our experiments, we could not establish a relationship between the degree of parallelism in the model and the generalization capability. However, our results do indicate that a process' complexity has an impact, but probably in more general terms (i.e. in terms of the number of variants). Given that we kept the number of variants contained, this should be confirmed with further research.

This work brings about a wide array of future research opportunities. With a distinct focus on next event prediction and a pure control-flow perspective, there is a clear opportunity for expansion towards other prediction tasks (outcome, remaining time, suffix), as well as towards the inclusion of additional variables such as resources or timestamps. Theoretical underpinnings of the generalization capacity of RNNs, in line with~\cite{Tu2020Understanding}, can also be a subject for further study. Furthermore, broadening the array of techniques included in the current assessment to convolutional neural networks (CNNs), Generative Adverserial Networks (GANs), and expanding the models to use different encoding techniques as in~\cite{Weinzierl_2020} seems worthwhile as well. Finally, the development of algorithmic and practical solutions to train RNN-based predictive process monitoring models in such a way that they attend to the common expectation that they should generalize process model structure, is a key future research challenge. Custom validation set creation, partial-order based data generation \cite{order}, as well as more finegrained overfitting countermeasures based on concepts like Layer Normalization~\cite{Ba_2016} and Recurrent Batch Normalization~\cite{Cooijmans_2016} are potential ideas for developing novel solutions. Other research possibilities might be found in exploring the (dis)advantage of setting aside information in the validation set. This could be done by investigating the balance between loss of information to train on and explicitly choosing your hyperparameters (and your early stopping) on new behavior. In this way a hyperparameter search similar to our post-hoc hyperparameter search could also be performed. An alternative approach would consist in combining both the strengths of algorithmic discovery techniques (as seen in Appendix~\ref{extra2}) and deep learning models, like done in~\cite{graphcombination}.

\subsubsection*{Competing interests}
The authors have no competing interests to declare that are relevant to the content of this article.

\subsubsection*{Availability of Data and Material}
The datasets generated during and/or analysed during the current study are available in the github repository, \url{https://github.com/jaripeeperkorn/LSTM_Process_Model_Structure}.

\bibliography{bibliography}


\newpage
\begin{appendices}

\section{Absolute Metrics}\label{extra}

Complementary to the metrics introduced in in Section \ref{Metrics}, we have designed three alternative metrics, which we call the ``absolute'' metrics and which can be seen below. These are not well calibrated, as they highly depend on the predetermined size of the \textit{Tr+Te} and the \textit{Simulated Log} (in this work always taken to be 100 times the number of variants). This is because the values depend on whether a certain variants present in another log at all, without taking multiplicities into account.For this we use the $Ex(v, L)$ function. Subsequently, $F_\textit{A-PMSL}$ corresponds with the fraction of variants present in the original \textit{Training Log} that are actually replicated in the \textit{Simulated Log}. $P_\textit{A-PMSL}$ then measures how many of the variants produced by the LSTM in the \textit{Simulated Log} actually corresponds with correct behavior, present in the original \textit{Tr+Te Log}. Finally, $G_\textit{A-PMSL}$ just counts how many of the variants in the \textit{Test Log} are correctly reproduced by the LSTM. In case of the LOVOCV $G_\textit{A-PMSL}$ is equal to either $0$ or $1$, since there is only one variant in the \textit{Simulated Log}. 

\begin{equation}
\label{eqn:absfitness}
    F_\textit{A-PMSL} = \sum_{v\in\textit{Var(Tr)}} \frac{Ex\left(v, Sim\right)}{\abs{\textit{Var(Tr)}}}
\end{equation}
\begin{equation}
\label{eqn:absprecision}
    P_\textit{A-PMSL} = \sum_{v\in\textit{Var(Sim)}} \frac{Ex\left(v, Tr+Te\right)}{\abs{\textit{Var(Sim)}}}
\end{equation}
\begin{equation}
\label{eqn:absgeneralization}
    G_\textit{A-PMSL} = \sum_{v\in\textit{Var(Te)}} \frac{Ex\left(v, Sim\right)}{\abs{\textit{Var(Te)}}}
\end{equation}
\begin{equation}
    \text{With: } \textit{Ex}(v, L) =
    \begin{cases}
  1 & \text{if } v \in L
\\
  0 & \text{if } v \not\in L
\end{cases}
\end{equation}

\begin{figure*}[ht!]
\captionsetup[subfigure]{labelformat=empty}
\begin{subfigure}{0.5\linewidth}
  \includegraphics[width=\linewidth]{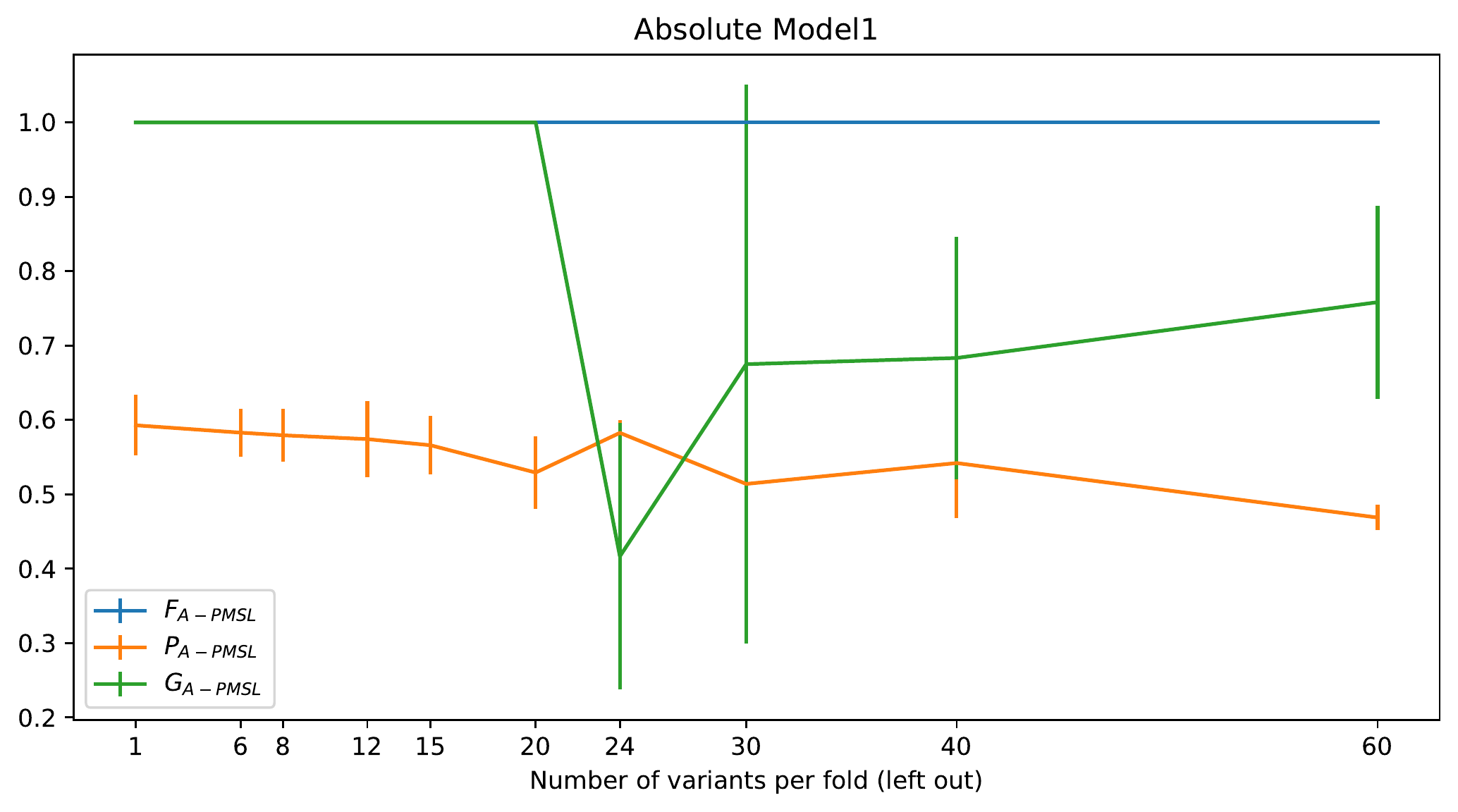}
\end{subfigure}
\hfill
\begin{subfigure}{0.5\linewidth}
  \includegraphics[width=\textwidth]{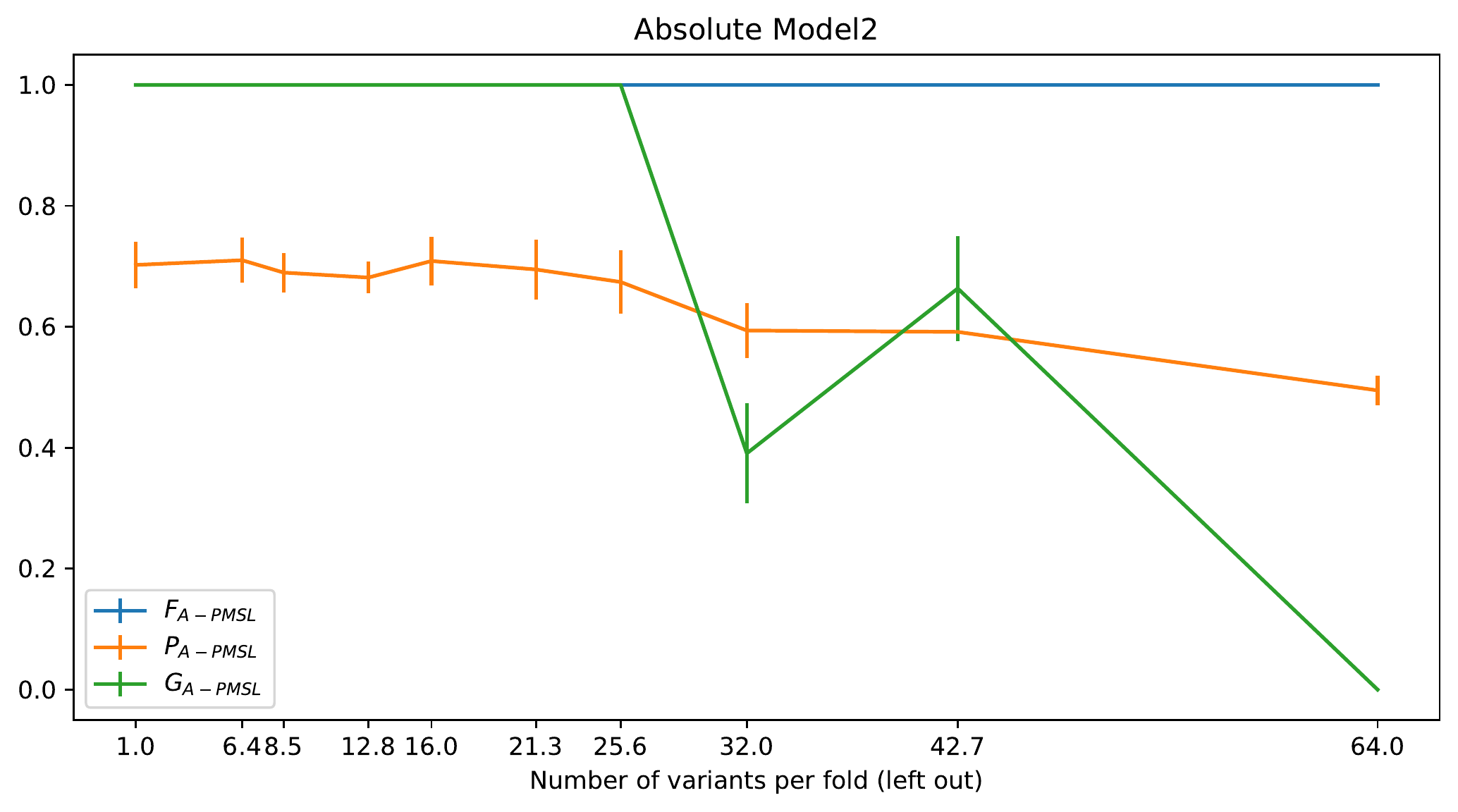}
\end{subfigure}
\hfill
\begin{subfigure}{0.5\linewidth}
  \includegraphics[width=\textwidth]{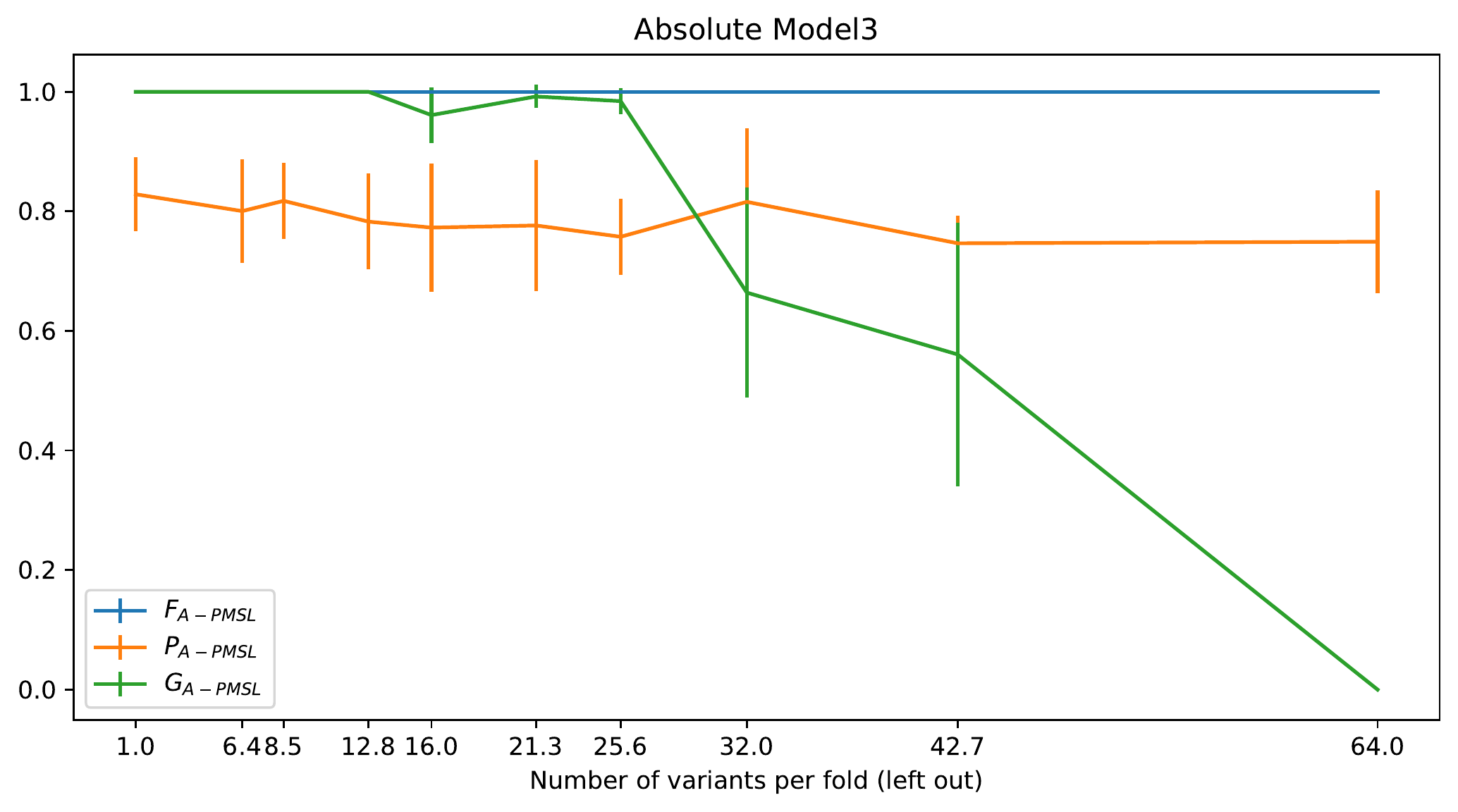}
\end{subfigure}
\hfill
\begin{subfigure}{0.5\linewidth}
  \includegraphics[width=\textwidth]{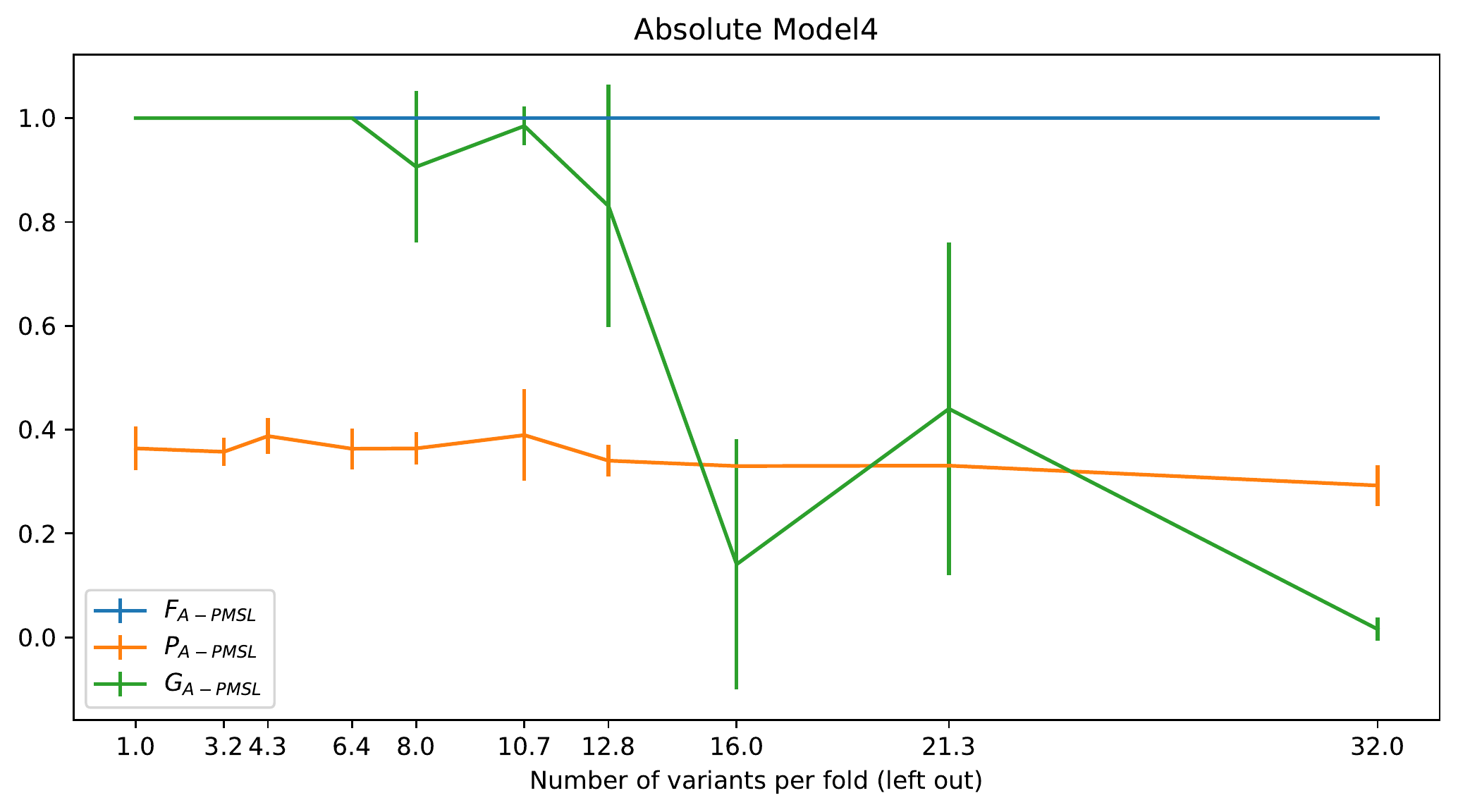}
\end{subfigure}
\hfill
\begin{subfigure}{0.5\linewidth}
  \includegraphics[width=\textwidth]{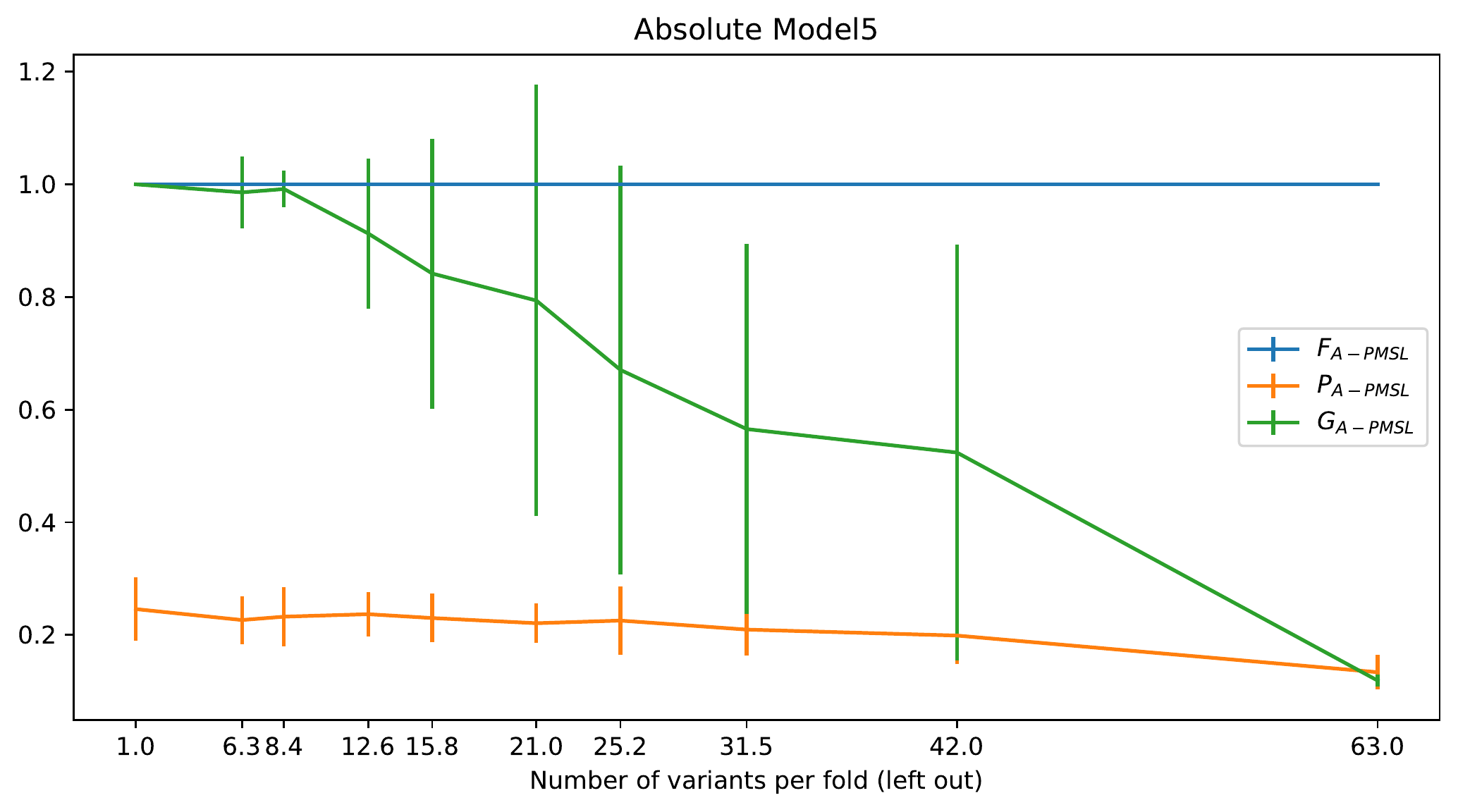}
\end{subfigure}
\hfill
\begin{subfigure}{0.5\linewidth}
\includegraphics[width=\textwidth]{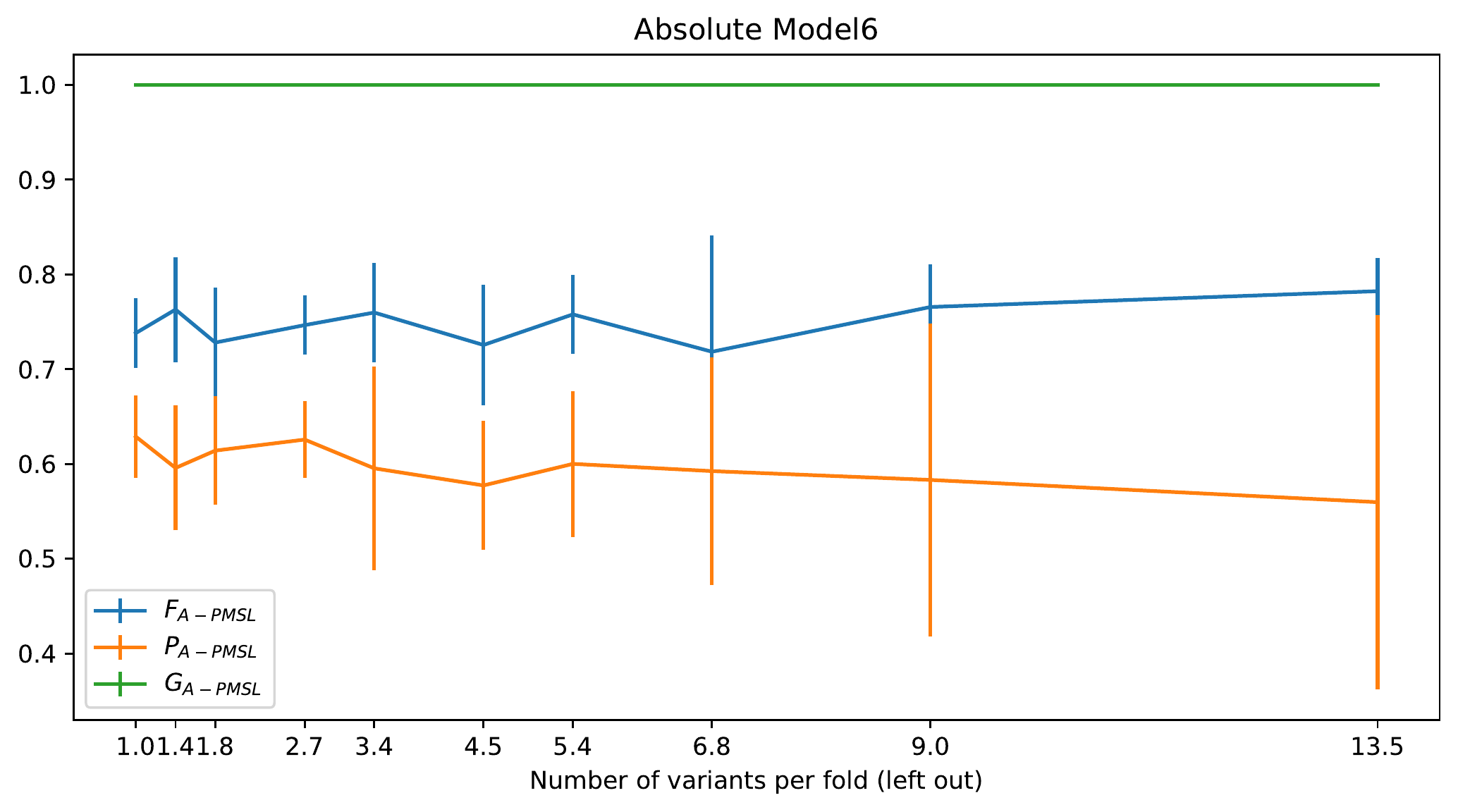}
\end{subfigure}
\hfill
\vspace{2mm}
 \caption{The framework output with the alternative absolute metrics as performed on LSTMs trained with different amounts of variants in the \textit{Test Log}. \label{Plots2}}
\label{fig:plotsabsolute}
\end{figure*}

Next to the metrics in the paper itself, these metrics have been applied on the logs, produced by the LSTMs discussed throughout (Table \ref{tab:Results3} and \ref{tab:Results4} and Figure \ref{fig:plotsabsolute}). Using these results we can unveil whether non-optimal results on the original metrics ($F_\textit{PMSL}$, $P_\textit{PMSL}$ and $G_\textit{PMSL}$) are due to the LSTM not learning certain behavior at all, or just because it displays lower probabilities to simulate these variants. The lower ``absolute'' precision scores ($P_\textit{A-PMSL}$), combined with the nonetheless high $P_\textit{PMSL}$ scores, shows that a lot of the LSTMs simulate a lot of different (wrong) variants, which do not correspond with allowed process model behavior; however do this with low multiplicities.

\begin{table}[ht]
    \centering
    \setlength\tabcolsep{3pt}
    \resizebox{\textwidth}{!}{
    \begin{tabular}{ccc|ccc|ccc}
    \hline
        &&& \multicolumn{3}{c|}{Acc. based Hyperparameters}
        & \multicolumn{3}{c}{Post-Hoc Hyperparameters}\\    
        \textbf{Mod.}  & \textbf{Pattern} & \textbf{\#Var.} & \textbf{$F_\textit{PMSL}$} & \textbf{$P_\textit{PMSL}$}  &\textbf{$G_\textit{PMSL}$} & \textbf{$F_\textit{PMSL}$} & \textbf{$P_\textit{PMSL}$}  &\textbf{$G_\textit{PMSL}$} \\ \hline
        1 & PAR & 120 
        & $1.00\pm0.00$ & $0.98\pm0.03$ & $0.16\pm0.37$ 
        & $1.00\pm0.00$ & $0.59\pm0.04$ & $1.00\pm0.00$ \\

        2 & XOR & 128
        & $1.00\pm0.00$ & $0.96\pm0.06$ & $1.00\pm0.00$ 
        & $1.00\pm0.00$ & $0.70\pm0.04$ & $1.00\pm0.00$ \\

        3 & XOR+LTD & 128
        & $1.00\pm0.00$ & $0.96\pm0.05$ & $0.20\pm0.40$ 
        & $1.00\pm0.00$ & $0.83\pm0.06$ & $1.00\pm0.00$ \\

        1 & IOR & 64
        & $1.00\pm0.00$ & $0.87\pm0.13$ & $1.00\pm0.00$ 
        & $1.00\pm0.00$ & $0.36\pm0.04$ & $1.00\pm0.00$ \\

       1 & PAR & 126
        & $1.00\pm0.00$ & $0.90\pm0.06$ & $0.37\pm0.48$ 
        & $1.00\pm0.00$ & $0.25\pm0.06$ & $1.00\pm0.00$ \\

        1 & LOOP & 27
        & $0.73\pm0.03$ & $0.68\pm0.04$ & $0.96\pm0.19$ 
        & $0.74\pm0.04$ & $0.63\pm0.04$ & $1.00\pm0.00$ \\
        \hline
    \end{tabular}}
    \caption{The results with the alternative absolute metrics on the different process models from Table \ref{tab:overview}, averaged over all leave-one-variant-out experiments with every different control flow variant. \label{tab:Results3}}
    \vspace{-10mm}
\end{table}

\begin{table}[h!]
    \centering
    \setlength\tabcolsep{3pt}
    \begin{tabular}{cc|ccc|ccc}
        \hline
        & & \multicolumn{3}{c|}{Acc. based Hyperparameters}
        & \multicolumn{3}{c}{Post-Hoc Hyperparameters}\\ 
        \textbf{Par.} & \textbf{\#Var.} & \textbf{$F_\textit{A-PMSL}$} & \textbf{$P_\textit{A-PMSL}$}  &\textbf{$G_\textit{A-PMSL}$} & \textbf{$F_\textit{A-PMSL}$} & \textbf{$P_\textit{A-PMSL}$}  &\textbf{$G_\textit{A-PMSL}$} \\ \hline
        0\% & 96 
        & $1.00\pm0.00$ & $0.86\pm0.11$ & $0.70\pm0.46$ 
        & $1.00\pm0.00$ & $0.78\pm0.08$ & $1.00\pm0.00$ \\ 
        5\% &108
        & $1.00\pm0.00$ & $0.86\pm0.10$ & $0.68\pm0.47$ 
        & $1.00\pm0.00$ & $0.46\pm0.09$ & $1.00\pm0.00$ \\ 
        10\% &120
        & $1.00\pm0.00$ & $0.92\pm0.08$ & $0.68\pm0.47$ 
        & $1.00\pm0.00$ & $0.57\pm0.06$ & $1.00\pm0.00$ \\ 
        15\% &150
        & $1.00\pm0.00$ & $0.93\pm0.08$ & $0.64\pm0.48$ 
        & $1.00\pm0.00$ & $0.46\pm0.06$ & $1.00\pm0.00$ \\ 
        20\% &136
        & $0.98\pm0.01$ & $0.93\pm0.08$ & $0.82\pm0.38$ 
        & $0.99\pm0.01$ & $0.56\pm0.06$ & $0.99\pm0.09$ \\ 
        25\% &108
        & $0.99\pm0.01$ & $0.85\pm0.11$ & $0.61\pm0.49$ 
        & $0.99\pm0.01$ & $0.32\pm0.03$ & $0.96\pm0.19$ \\ 
        30\% &160
        & $1.00\pm0.00$ & $0.94\pm0.06$ & $0.58\pm0.49$ 
        & $1.00\pm0.00$ & $0.36\pm0.02$ & $1.00\pm0.00$ \\ 
        35\% &140
        & $1.00\pm0.00$ & $0.95\pm0.05$ & $0.53\pm0.50$ 
        & $1.00\pm0.00$ & $0.39\pm0.05$ & $1.00\pm0.00$ \\ 
        40\% &120
        & $1.00\pm0.00$ & $0.91\pm0.08$ & $0.57\pm0.50$ 
        & $1.00\pm0.00$ & $0.50\pm0.05$ & $1.00\pm0.00$ \\ 
        45\% &120
        & $1.00\pm0.00$ & $0.93\pm0.08$ & $0.71\pm0.50$ 
        & $1.00\pm0.00$ & $0.60\pm0.06$ & $1.00\pm0.00$ \\ 
        50\% &144
        & $1.00\pm0.00$ & $0.91\pm0.10$ & $0.54\pm0.46$ 
        & $1.00\pm0.00$ & $0.37\pm0.04$ & $1.00\pm0.00$ \\ \hline
    \end{tabular}
    \caption{The results with the alternative absolute metrics on the different process models from Table \ref{tab:overview2}, averaged over all leave-one-variant-out experiments with every different control flow variant. \label{tab:Results4}}
    \vspace{-6mm}
\end{table}

\newpage

\section{Comparing with Inductive Miner}\label{extra2}
\renewcommand{\thefigure}{B\arabic{figure}}
\setcounter{figure}{0}

\begin{figure*}[!htb]
\captionsetup[subfigure]{labelformat=empty}
\begin{subfigure}{0.5\linewidth}
  \includegraphics[width=\linewidth]{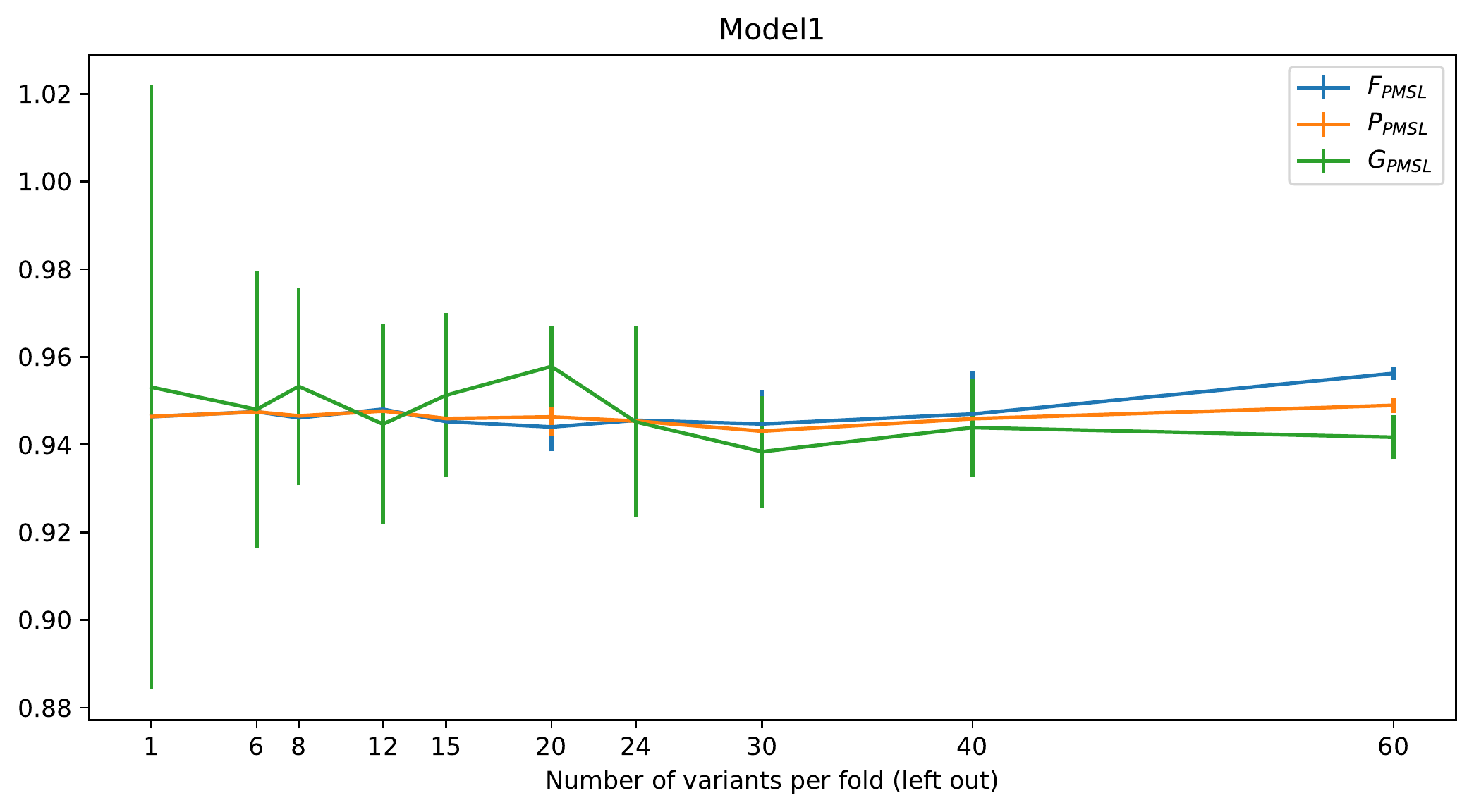}
\end{subfigure}
\hfill
\begin{subfigure}{0.5\linewidth}
  \includegraphics[width=\textwidth]{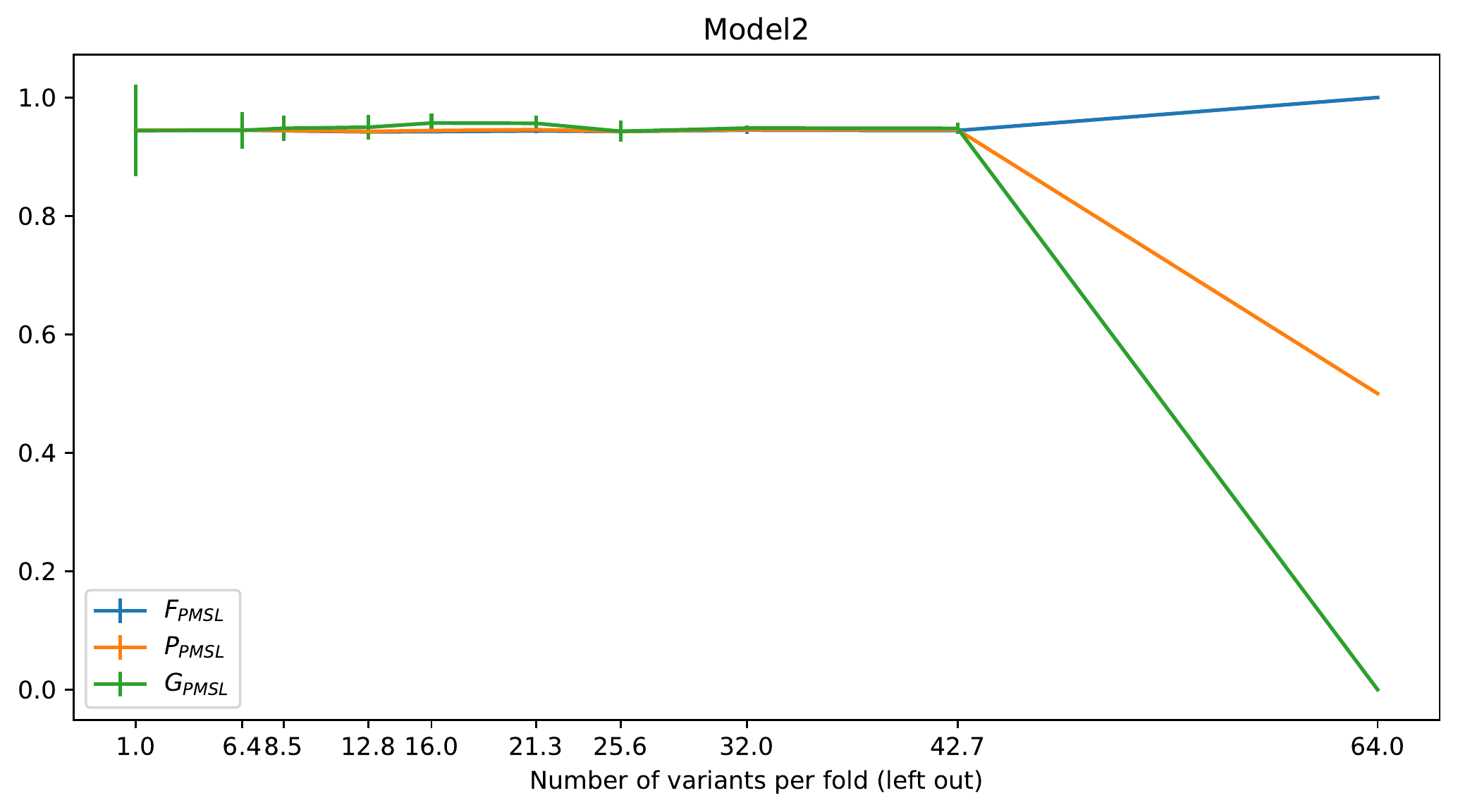}
\end{subfigure}
\hfill
\begin{subfigure}{0.5\linewidth}
  \includegraphics[width=\textwidth]{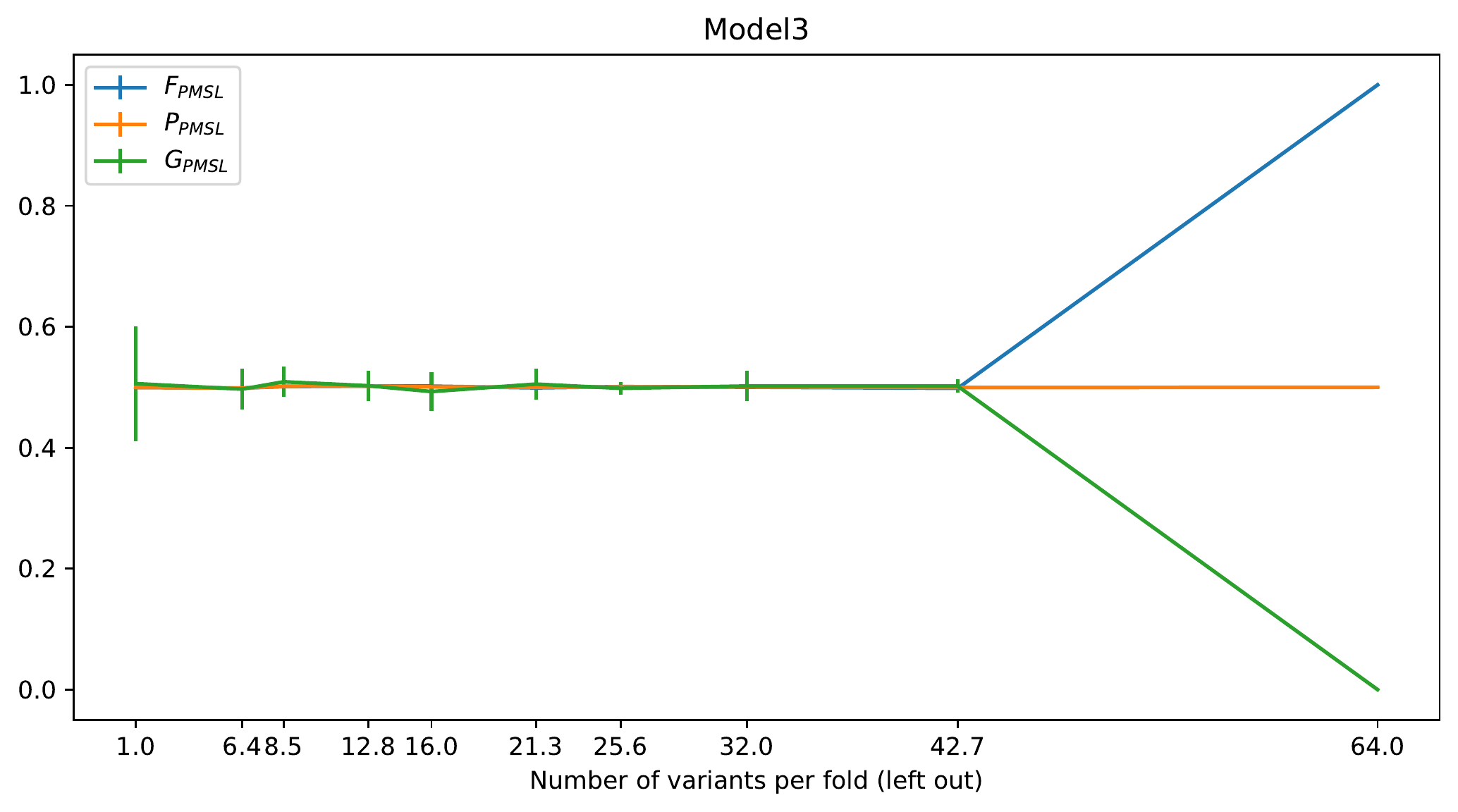}
\end{subfigure}
\hfill
\begin{subfigure}{0.5\linewidth}
  \includegraphics[width=\textwidth]{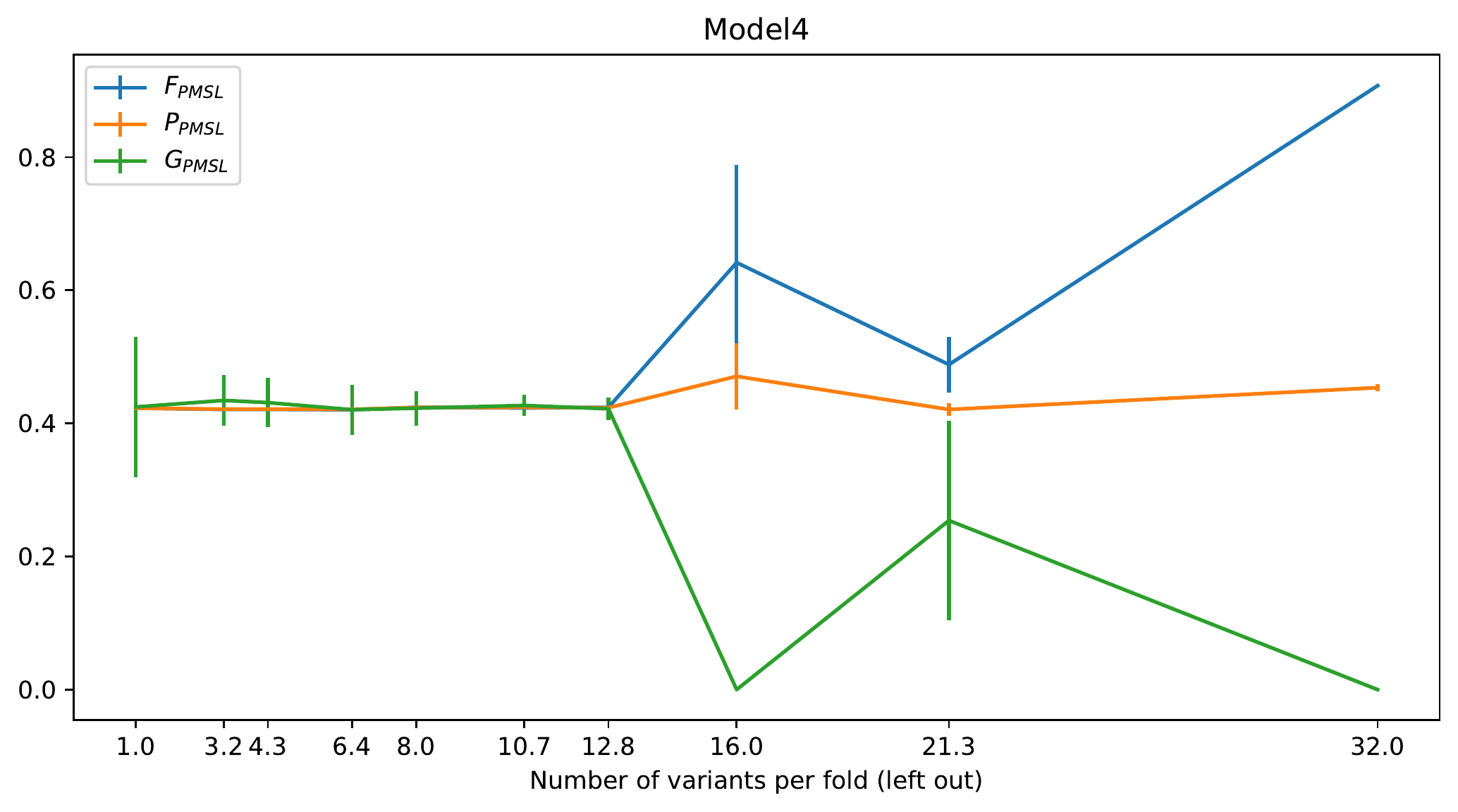}
\end{subfigure}
\hfill
\begin{subfigure}{0.5\linewidth}
  \includegraphics[width=\textwidth]{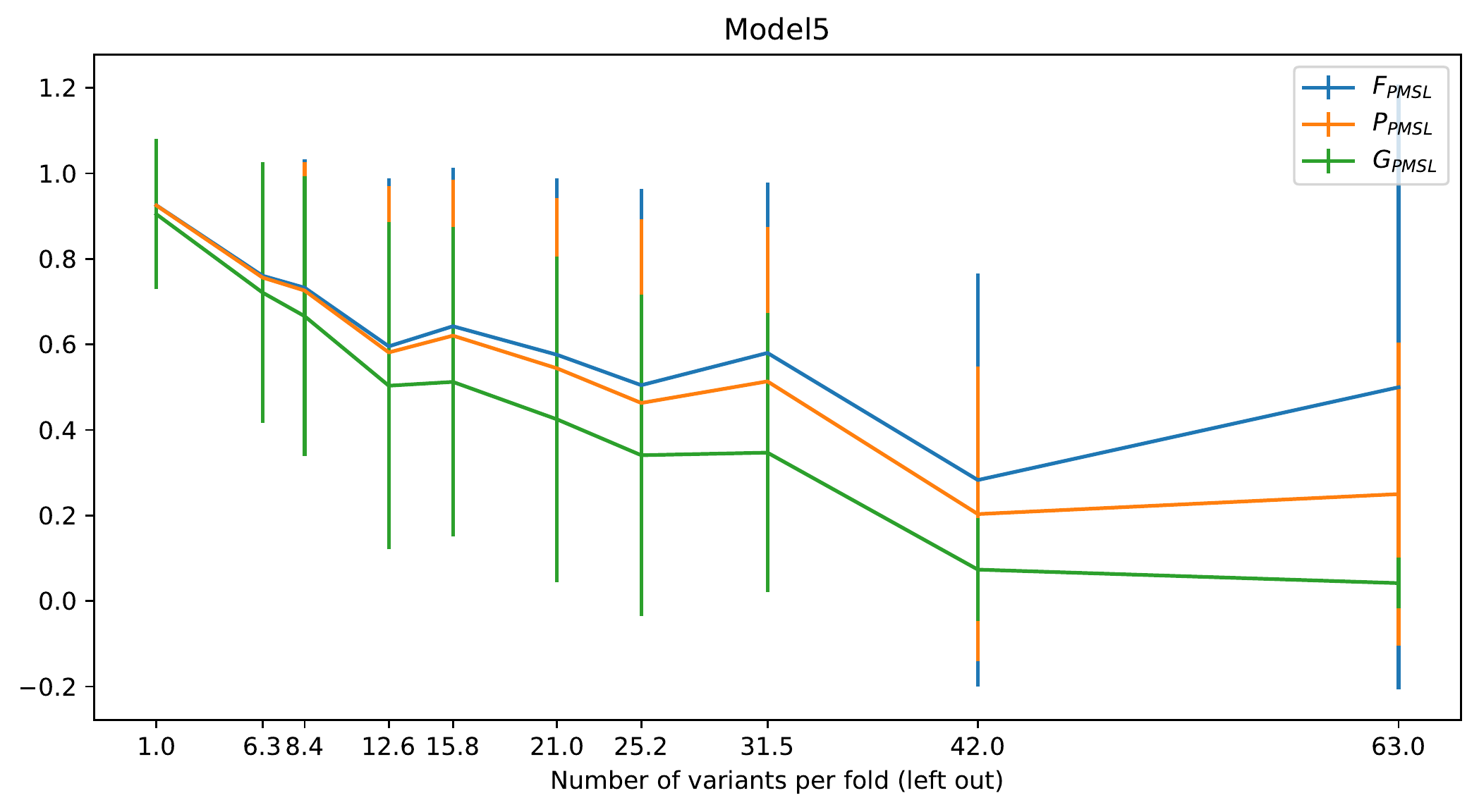}
\end{subfigure}
\hfill
\begin{subfigure}{0.5\linewidth}
\includegraphics[width=\textwidth]{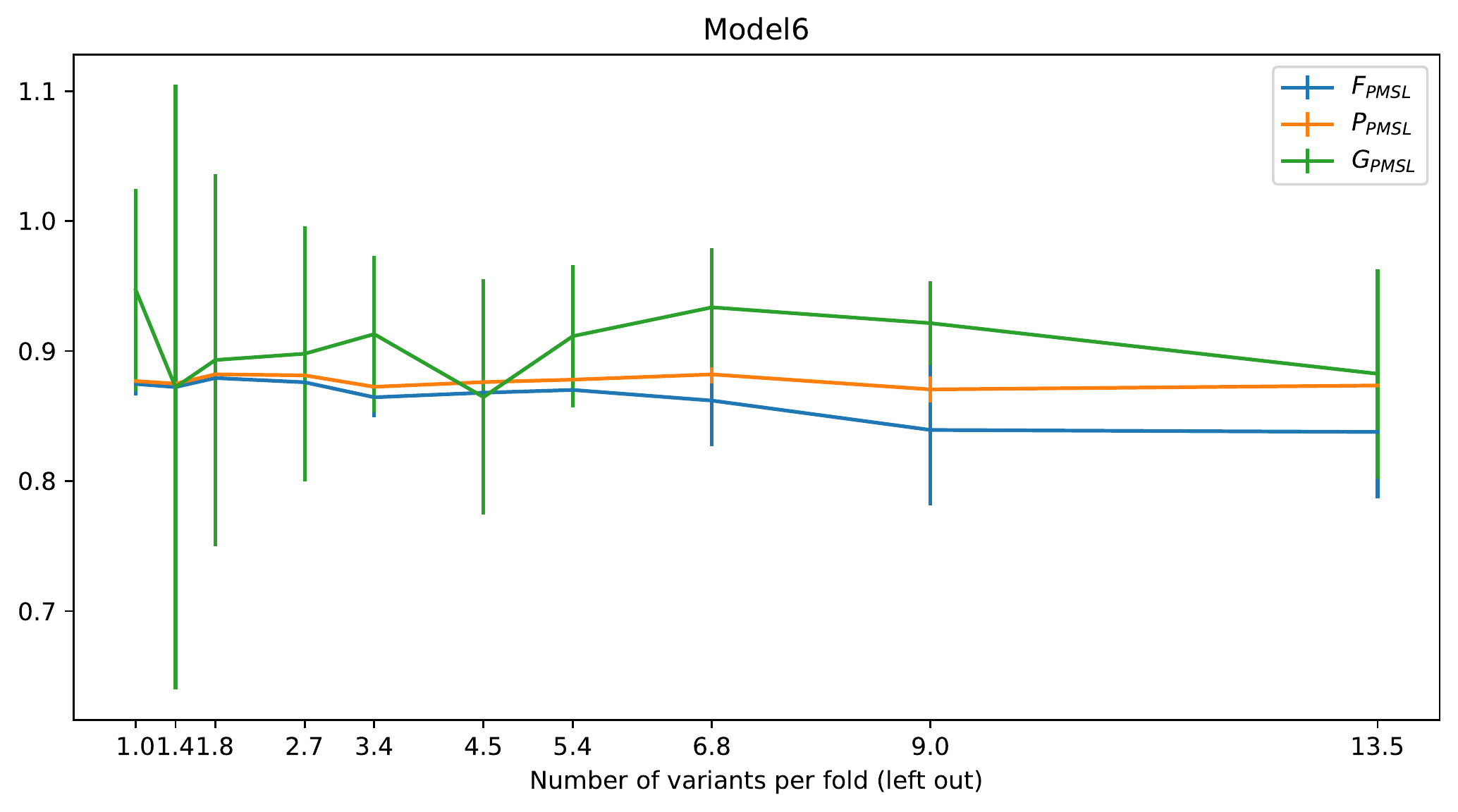}
\end{subfigure}
\hfill
\vspace{2mm}
 \caption{The framework output, using metrics~\ref{eqn:fitness}-\ref{eqn:generalization}, as performed on the simulated Logs produced by playing out Process Trees discovered with the Inductive Miner~\cite{inductive}, discovered with different amounts of variants left out and put aside in the \textit{Test Log}.}
\label{fig:plotsinductive}
\vspace{-6mm}
\end{figure*}

\begin{figure*}[!htb]
\captionsetup[subfigure]{labelformat=empty}
\begin{subfigure}{0.5\linewidth}
  \includegraphics[width=\linewidth]{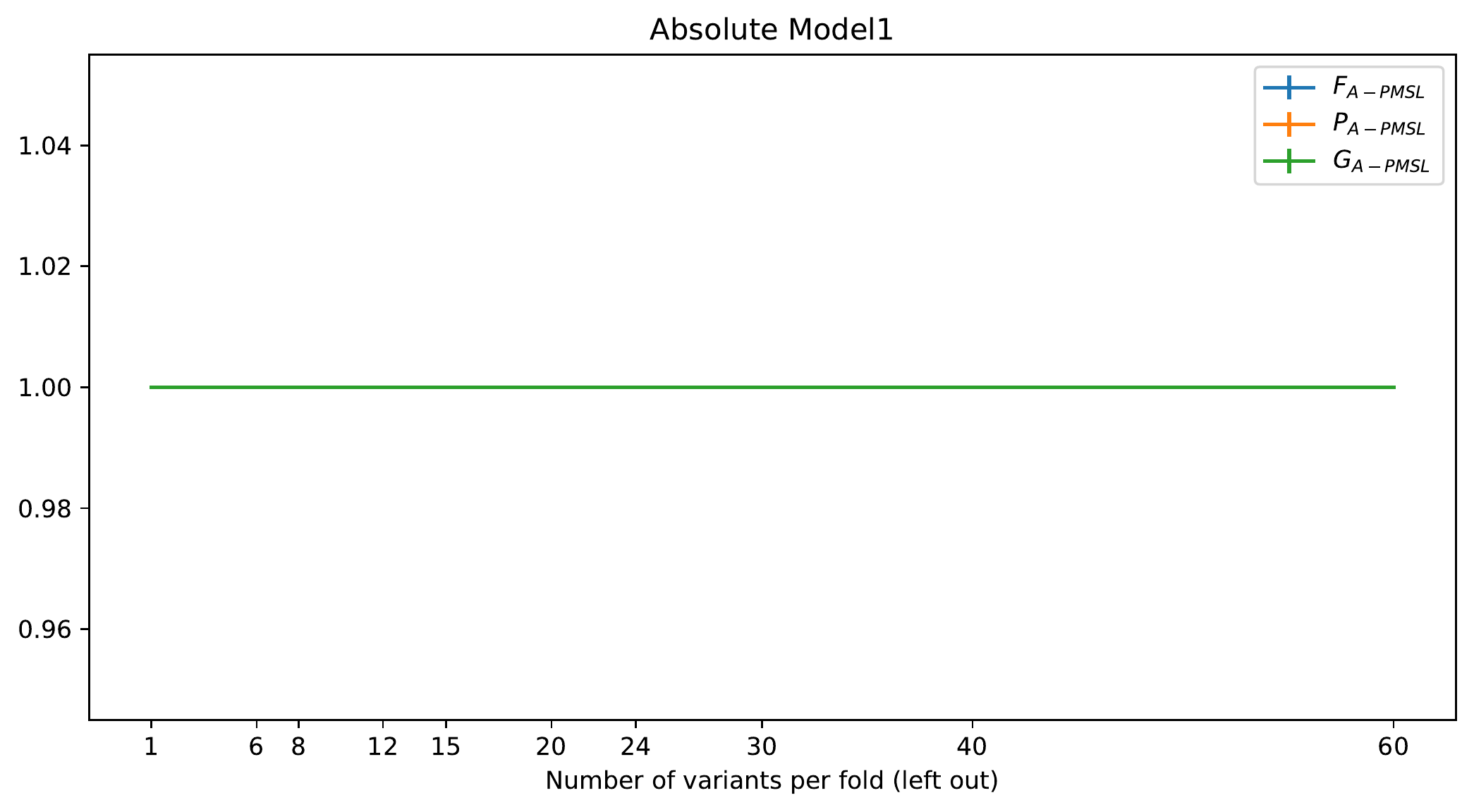}
\end{subfigure}
\hfill
\begin{subfigure}{0.5\linewidth}
  \includegraphics[width=\textwidth]{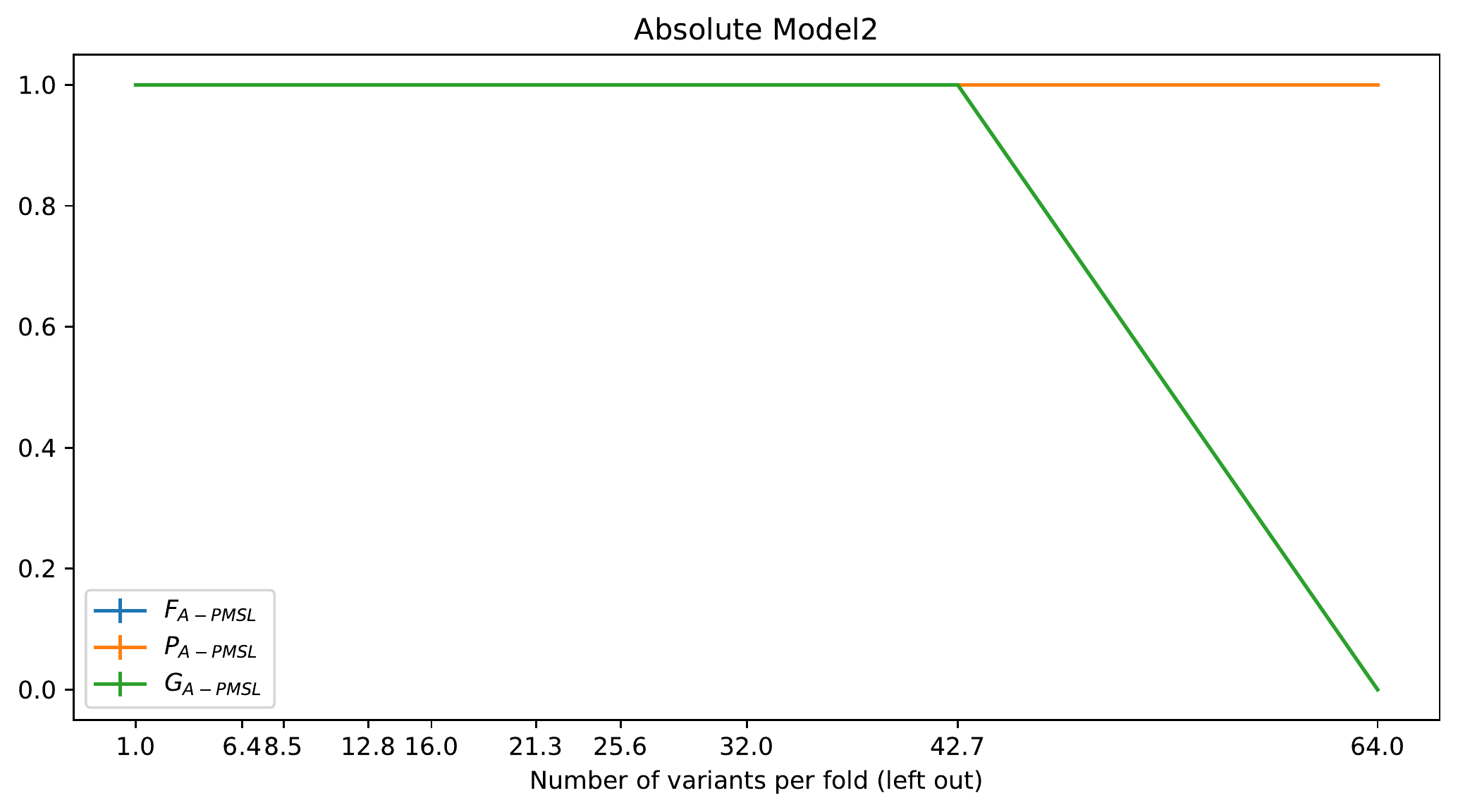}
\end{subfigure}
\hfill
\begin{subfigure}{0.5\linewidth}
  \includegraphics[width=\textwidth]{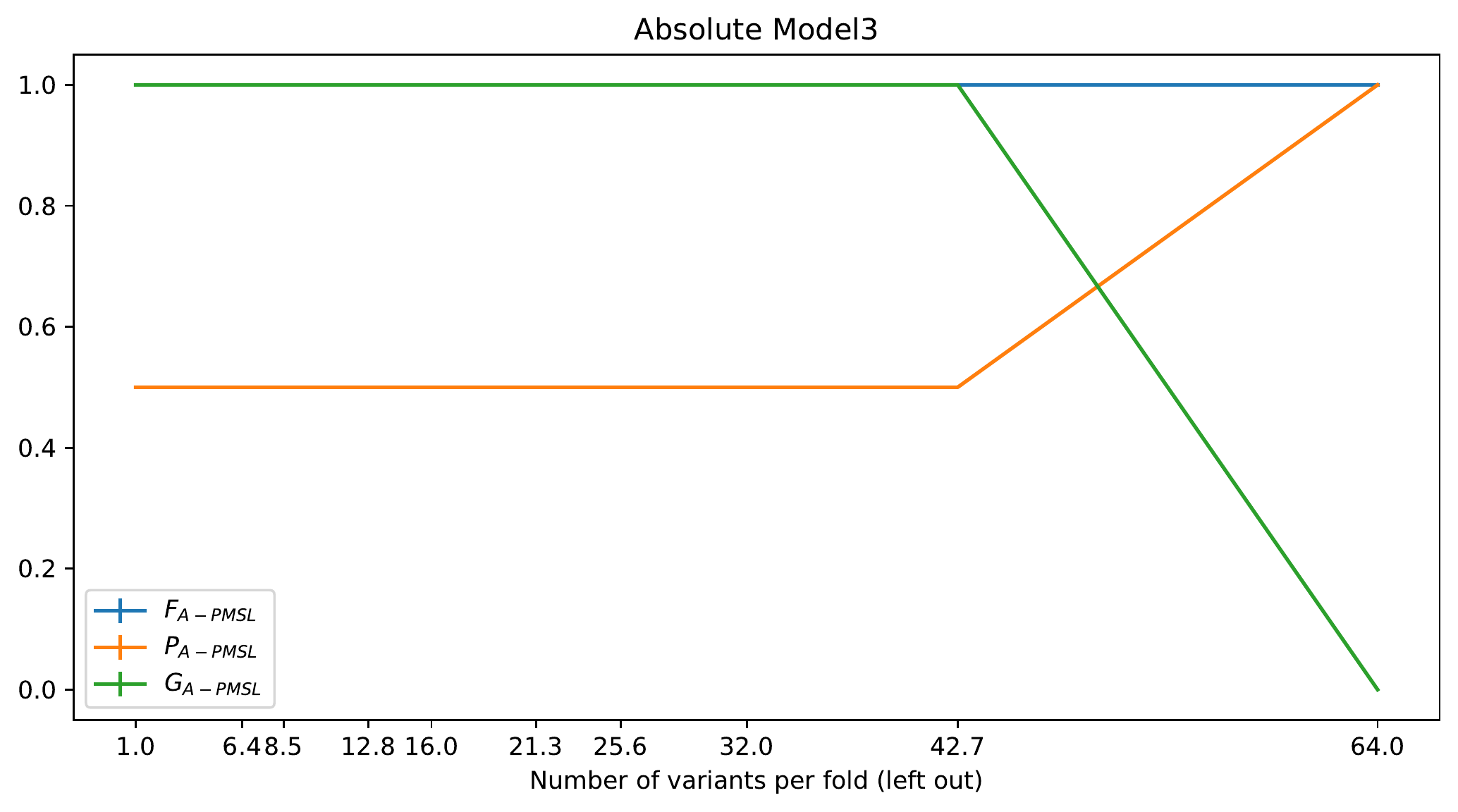}
\end{subfigure}
\hfill
\begin{subfigure}{0.5\linewidth}
  \includegraphics[width=\textwidth]{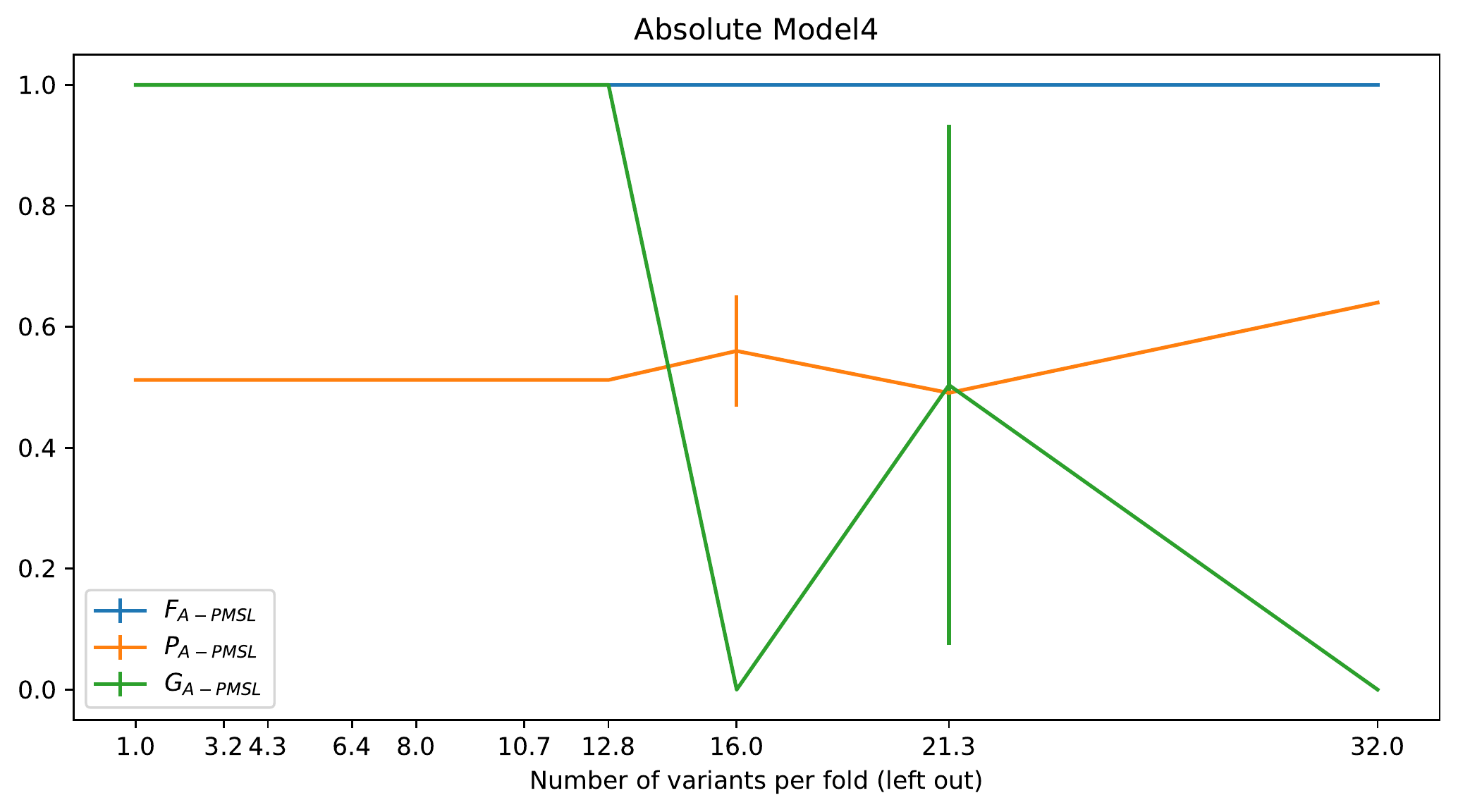}
\end{subfigure}
\hfill
\begin{subfigure}{0.5\linewidth}
  \includegraphics[width=\textwidth]{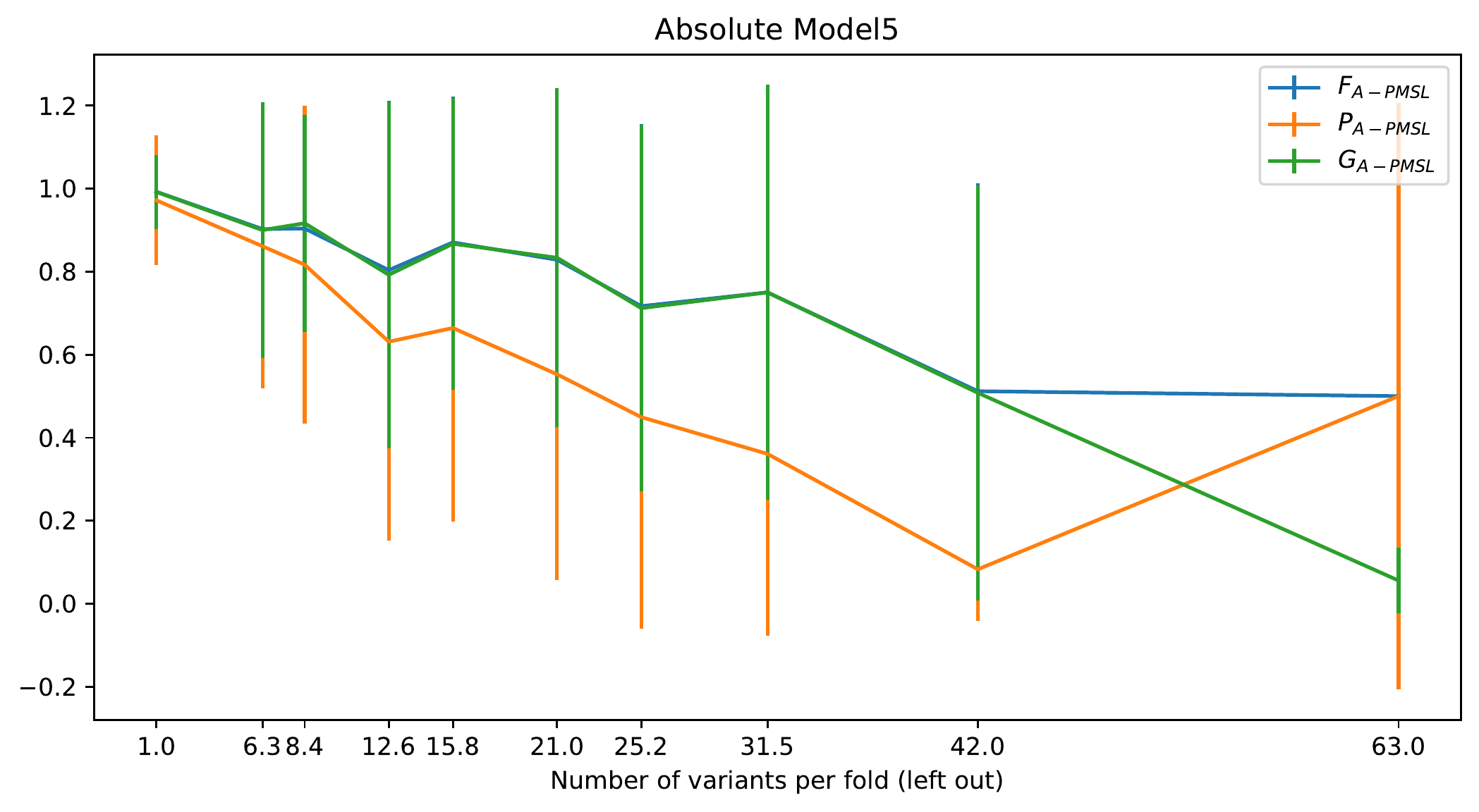}
\end{subfigure}
\hfill
\begin{subfigure}{0.5\linewidth}
\includegraphics[width=\textwidth]{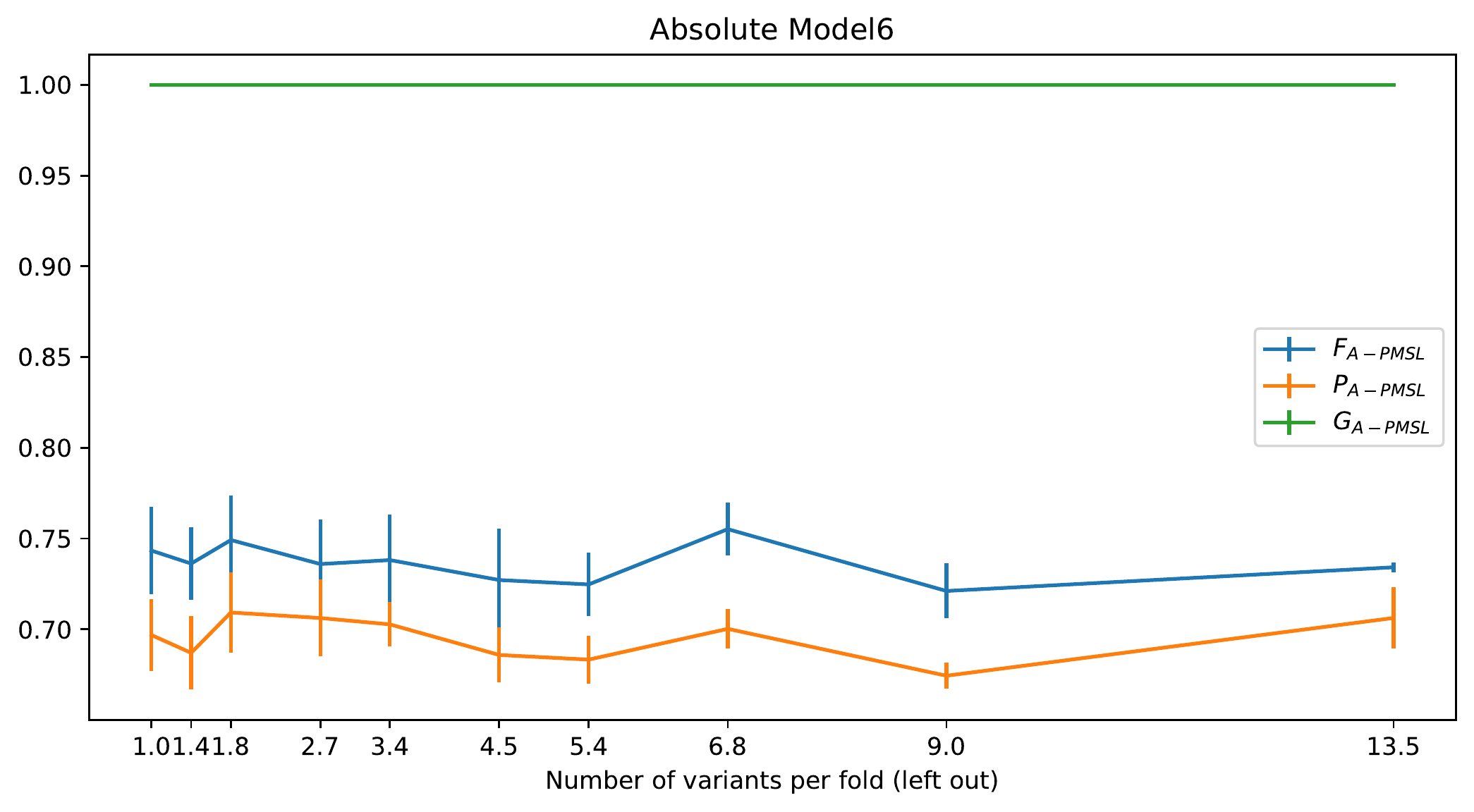}
\end{subfigure}
\hfill
\vspace{2mm}
 \caption{The framework output, using the alternative absolute metrics~\ref{eqn:absfitness}-\ref{eqn:absgeneralization}, as performed on the simulated Logs produced by playing out Process Trees discovered with the Inductive Miner~\cite{inductive}, discovered with different amounts of variants left out and put aside in the \textit{Test Log}.}
\label{fig:plotsinductiveabsolute}
\vspace{-6mm}
\end{figure*}

To compare the LSTMs' abilities to understand the six different basic control flow elements introduced in Figure~\ref{fig:models} to ``classical'' process discovery, we have performed some additional experiments. For this purpose we have used the Inductive Miner~\cite{inductive} as implemented in pm4py~\cite{pm4py}, with the default settings (corresponding to a noise threshold of 0.0) to discover Process Trees. These process Trees are further played out (simulated) with the basic play out function, implemented in pm4py~\cite{pm4py} as well. We have repeated both the LOVOCV and the fold experiment introduced in Chapter~\ref{Setup}. In the fold experiment we take out an increasing number of variants from the \textit{Training Log} to group them into the \textit{Test Log}. We have used both the metrics from Equations~\ref{eqn:fitness}-\ref{eqn:generalization} and the ``absolute'' metrics~\ref{eqn:absfitness}-\ref{eqn:absgeneralization} introduced in Appendix A. The results of these experiments are visualized in Figures~\ref{fig:plotsinductive} and ~\ref{fig:plotsinductiveabsolute}. We can immediately notice that Model 1, the model with one parallel split, is discovered (and played out)  perfectly by the Inductive Miner, and this for both the LOVOCV as for every fold. This is in contrast to the LSTM models, which seem to struggle a bit more with correctly interpreting this kind of behavior. Model 2, introducing multiple XOR splits, also does not cause any issues for the Inductive Miner, with the exception of the 2 fold experiment (where half of the control-flow variants are sampled out of the \textit{Training Log}). The long-term dependency introduced in Model 3 is not picked up at all by the Process Tree discovered with the Inductive Miner (unsurprisingly). This results in near perfect absolute fitness~\ref{eqn:absfitness} and generalization~\ref{eqn:absgeneralization} scores (the Process Tree is able to play out the behavior), but frequency dependent fitness~\ref{eqn:fitness} and generalization~\ref{eqn:generalization} scores of around $0.5$. Both precision scores are also around $0.5$. Remarkably when performing the 2 fold experiment, fitness (and absolute precision) goes to $1.0$ and generalization to $0.0$. What happens is that the model completely overfits on the possible control-flow variants (one big XOR split) preventing any level of generalization. A similar type of behavior can be noticed for Model 4, which consists out of different inclusive OR splits. The longer parallel tracks of Model 5 are interpreted worse and worse, as more and more variants are left out of the \textit{Training Log}, leading to a decreasing trend in all metrics. This is comparable to the results of the LSTMs trained on the same model. It is in contrast to the better interpretation of the Inductive Miner of the parallel split in Model 1 (with short 1 activity long parallel tracks).  Finally the loops in Model 6 are modeled correctly by the Inductive Miner, where the imperfect frequency based fitness and precision scores are due to the actually infinite amount of possible behavior introduced by loops. 

\end{appendices}

\end{document}